\documentclass{article}
\usepackage{iclr2027_conference,times}

\usepackage{microtype}
\usepackage{graphicx}
\usepackage{subcaption}
\usepackage{wrapfig}
\usepackage{booktabs}
\usepackage{xcolor}

\usepackage{amsmath}
\usepackage{amssymb}
\usepackage{mathtools}
\usepackage{amsthm}
\usepackage{graphicx}
\usepackage{multirow}
\usepackage{multicol}

\usepackage{xstring}
\usepackage{tikz}
\usetikzlibrary{calc}

\usepackage{amsmath,amsfonts,bm}

\def\eqref#1{equation~\ref{#1}}

\def\1{\bm{1}}

\def\vzero{{\bm{0}}}

\def\vtheta{{\bm{\theta}}}
\def\va{{\bm{a}}}
\def\vb{{\bm{b}}}
\def\vc{{\bm{c}}}
\def\vd{{\bm{d}}}
\def\ve{{\bm{e}}}

\def\vh{{\bm{h}}}

\def\vk{{\bm{k}}}

\def\vo{{\bm{o}}}
\def\vp{{\bm{p}}}
\def\vq{{\bm{q}}}
\def\vr{{\bm{r}}}

\def\vu{{\bm{u}}}
\def\vv{{\bm{v}}}

\def\vx{{\bm{x}}}
\def\vy{{\bm{y}}}
\def\vz{{\bm{z}}}

\def\mA{{\bm{A}}}
\def\mB{{\bm{B}}}
\def\mC{{\bm{C}}}

\def\mE{{\bm{E}}}

\def\mH{{\bm{H}}}
\def\mI{{\bm{I}}}

\def\mK{{\bm{K}}}

\def\mW{{\bm{W}}}

\def\mZ{{\bm{Z}}}

\DeclareMathAlphabet{\mathsfit}{\encodingdefault}{\sfdefault}{m}{sl}
\SetMathAlphabet{\mathsfit}{bold}{\encodingdefault}{\sfdefault}{bx}{n}

\def\gC{{\mathcal{C}}}

\def\gE{{\mathcal{E}}}

\def\gL{{\mathcal{L}}}

\def\gN{{\mathcal{N}}}
\def\gO{{\mathcal{O}}}

\def\gU{{\mathcal{U}}}

\def\gX{{\mathcal{X}}}

\newcommand{\E}{\mathbb{E}}
\newcommand{\Ls}{\mathcal{L}}
\newcommand{\R}{\mathbb{R}}

\newcommand{\softmax}{\mathrm{softmax}}

\newcommand{\Var}{\mathrm{Var}}

\usepackage{hyperref}
\usepackage{url}

\usepackage{pifont}

\usepackage[capitalize,noabbrev]{cleveref}

\newif\ifdev
\devfalse

\usepackage{multirow}
\usepackage{soul}
\usepackage{xspace}

\newcommand{\rulecentreshift}[1]{%
    \dimexpr(\belowrulesep+#1-\aboverulesep)/2\relax}

\newcommand{\datasetlabel}[1]{%
    \rotatebox[origin=c]{90}{\normalfont\itshape\small #1}}

\newcommand{\methodsperblock}{5}

\newcommand{\blockspaceabove}{0pt}

\newlength{\dsrowstep}\newlength{\dsgapstep}
\AtBeginDocument{%
    \sbox0{\begin{tabular}{c}a\\ \end{tabular}}%
    \sbox2{\begin{tabular}{c}a\\ a\\ \end{tabular}}%
    \setlength{\dsrowstep}{\dimexpr\ht2+\dp2-\ht0-\dp0\relax}%
    \sbox4{\begin{tabular}{c}a\\ \cmidrule(lr){1-1}\addlinespace[2pt] a\\ \end{tabular}}%
    \setlength{\dsgapstep}{\dimexpr\ht4+\dp4-\ht2-\dp2\relax}%
}

\newcommand{\dataset}[3][0]{%
    \multirow{\the\numexpr\methodsperblock*#3\relax}{*}%
    [\dimexpr-\dsgapstep*\numexpr2*#3-1\relax/2+\rulecentreshift{\blockspaceabove}
    -(\methodsperblock\dsrowstep+2\dsgapstep)*#1/2\relax]%
    {\datasetlabel{#2}}}

\newcommand{\metric}[1]{{\scriptsize #1}}

\newcommand{\winning}[1]{\textbf{#1}}

\setul{1.2pt}{0.5pt}

\definecolor{unsurebg}{RGB}{255,214,214}
\sethlcolor{unsurebg}
\ifdev
    
\else
    
\fi

\definecolor{ongoingbg}{RGB}{150,201,150}
\ifdev
    
\else
    
\fi

\ifdev
    \newcommand{\strike}[1]{\st{#1}}
\else
    \newcommand{\strike}[1]{}
\fi

\newcommand{\veps}{{\bm{\epsilon}}}
\newcommand{\valpha}{{\bm{\alpha}}}

\newcommand{\method}{\textsc{Smart}}

\theoremstyle{plain}

\theoremstyle{definition}

\theoremstyle{remark}

\ifdev
\usepackage[textsize=tiny]{todonotes}
\else
\usepackage[disable,textsize=tiny]{todonotes}
\fi

\title{Long-horizon Video Generation with \\Slot Memory Addressed By Region Tokens}

\author{%
Eren Cetin$^{1}$ \quad Julius Erbach$^{1}$ \quad Vukasin Bozic$^{1}$ \quad Hyojun Go$^{1,2}$ \quad Zixiang Zhao$^{1}$ \\
\bfseries Nikolai Kalischek$^{2}$ \quad Federico Tombari$^{2}$ \quad Konrad Schindler$^{1}$ \quad Dominik Narnhofer$^{1}$ \\[4pt]
{\normalfont $^{1}$ETH Zurich \qquad $^{2}$Google}
}

\iclrfinalcopy
\begin{document}

\maketitle
\lhead{Preprint}

\begin{abstract}
Memory for long-horizon video generation must satisfy two competing requirements: it must remain bounded so generation can continue indefinitely, while still recovering what was observed at a place when the camera returns, regardless of how long ago it was seen. Reconstruction-based memories provide spatial recall by caching and reprojecting explicit geometry, but require growing storage and depend on estimated 3D structure. Recurrent memories keep a constant-size state, but retention is governed by temporal recurrence, causing old observations to fade with time. We introduce \method{}, a \underline{s}lot \underline{m}emory \underline{a}ddressed by \underline{r}egion \underline{t}okens, which combines bounded storage with spatially addressed recall. \method{} stores content in a constant-size slot recurrence whose writes are addressed by the observing camera and whose reads are queried by the rendering camera, without constructing an explicit 3D representation or maintaining a growing cache. Consequently, recall is conditioned on inter-view geometry rather than on recency. Each read additionally provides a confidence estimate that conditions a chunk-causal diffusion transformer, allowing it to selectively trust retrieved content while relying on its generative prior elsewhere. On long trajectories in Memory Maze and Minecraft, \method{} maintains high-fidelity recall far beyond its training horizon while using constant memory and per-step cost.%
\end{abstract}

\begin{figure}[h]
    \centering
    \vspace{-.9em}
    \includegraphics[width=.85\linewidth]{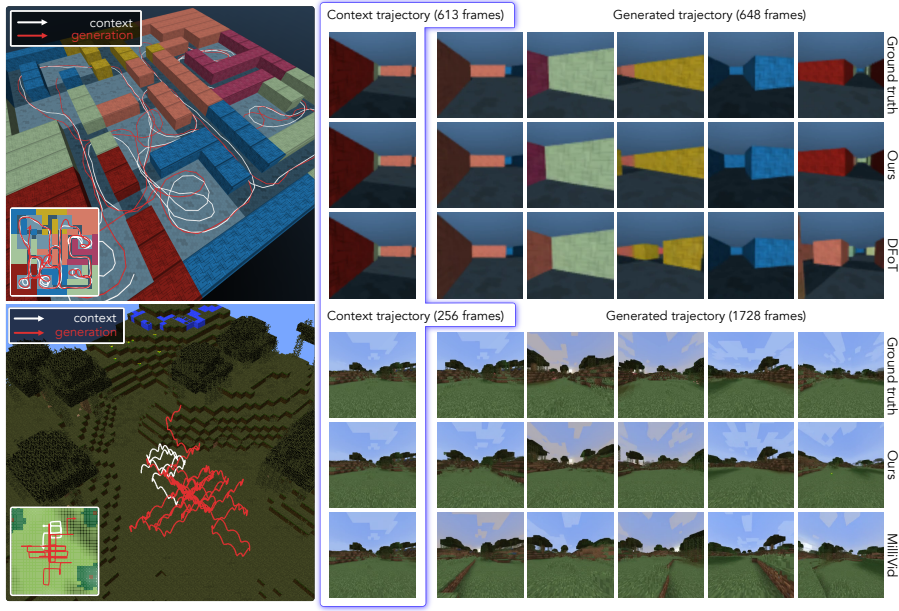}
    \vspace{-1.0em}
    \caption{\textbf{Long-horizon video generation.}
    \textit{Top:} on \textit{Memory Maze}, \method{} and DFoT~\citep{song2025historyguided} observe a forward trajectory and generate the return leg, revisiting places seen long before.
    \textit{Bottom:} on \textit{Minecraft}, \method{} and MilliVid~\citep{chandratreya2026millivid} generate a continuation trajectory from $256$ context frames written into the memory.
    \method{} retains observed content more faithfully than the baselines at constant per-step cost.}
    \label{fig:teaser}
\end{figure}

\section{Introduction}

{

Video generative models have improved dramatically in recent years. Large scale training and autoregressive generation have enabled models to produce increasingly long videos while following camera and action controls with high fidelity. However, even compressed latent videos consist of thousands of tokens, making it costly to train and condition on long sequences. Moreover, a generative model must implicitly learn complex long-range dependencies, such as those imposed by multi-view consistency. Even large-scale autoregressive models~\citep{bahmani2026lyra,longlive2,chen2024diffusion} struggle with this: when an agent walks down a corridor and subsequently retraces its path, the environment may no longer match its initial state. Remembering scene geometry over long horizons therefore remains difficult even at scale. Conditioning on a persistent representation of previously observed content can alleviate both problems. Rather than requiring the model to retain the full history and reconstruct cross-view relationships implicitly, such a representation can summarize past observations and provide information relevant to the current viewpoint. The central question thus becomes, how this representation should be structured.

Two families of methods have emerged to equip video generation with memory, which differ in terms of how memory is addressed. 
\emph{Explicit} memory directly instantiates the underlying 3d geometry in the form of Gaussians, point clouds or a latent 3D map, and reprojects it to the target camera to condition the generator~\citep{bahmani2026lyra,shen2026lyra,mosaicmem2026,wei2026gimworld,zhao2026spatia,wang2026mirage}.
The memory is addressed by a spatial location: content can be retrieved according to where it was observed. Explicit methods, therefore, enforce some level of long-term consistency by construction. However, the required storage generally grows with the size of the explored scene unless information is explicitly pruned, and the resulting rendered conditioning provides the generator only with limited information about which reconstructed regions are reliable.
\emph{Implicit} memory, in the form of linear-time recurrent sequence models, stores the memory within the neural architecture~\citep{Po_2025_ICCV,savov2026statespacediffuser}: in language models, state-space approaches~\citep{s4,mamba,mamba2}, gated linear attention and slot recurrences~\citep{gsa,mom_mixtureofmemories,raven2026} all compress an unbounded history into a constant-size state, evaluated on exactly the retrieval regime that concerns us. However, such memory schemes lack an explicit spatial addresses: information is accumulated through temporal recurrence rather than stored according to where it was observed. Moreover, every frame updates the entire state and decays older information, so what is retained  inherently depends on recency rather than spatial location. Old observations beyond the training horizon inevitably fade out.

We propose a conditioning mechanism that combines the complementary strengths of both approaches: the bounded, learned representation of implicit memory and the spatial addressing of explicit geometry.  We introduce \method{} (\textbf{s}lot \textbf{m}emory \textbf{a}ddressed by \textbf{r}egion \textbf{t}okens), which couples a fixed-capacity implicit memory, in which camera pose determines where observations are written and from where they are retrieved, with a video diffusion model conditioned on what that memory returns. Rather than asking the model to recover an observation from a particular (past) time step, a query decodes the memory into a coarse representation of the target view together with a confidence estimate. This provides the generative model with a full-frame conditioning signal, preserving the convenient interface of explicit reconstruction-based methods; but without the requirement to keep a full 3D scene reconstruction in memory.
Our contributions are:
\begin{itemize}
    \item A fixed-capacity slot-based implicit scene memory that addresses observations by where rather than when they were seen, enabling geometry-guided writing, retrieval, and novel-view decoding.
    \item A memory-conditioned video diffusion model that takes the retrieved view as a coarse draft and refines it into video, falling back on its generative prior where memory is uninformative.
\end{itemize}
}

\section{Related Work}\label{sec:related}
{
\noindent\textbf{Video diffusion and controllable generation.}
Diffusion models have become a strong foundation for video generation, with large-scale architectures such as Wan~\citep{wan2025} providing high-quality spatiotemporal synthesis. Camera-controlled generation further conditions on viewing geometry, using Pl\"ucker-ray representations~\citep{He_2025_ICCV}, geometry-aware attention~\citep{Miyato2024GTA}, or relative-camera positional encodings~\citep{li2025cameras,rerope2026siggraph}. These mechanisms condition the current step on pose, but leave open what the model retains from earlier steps.\\
\noindent\textbf{Temporal context for video generation.} Autoregressive and causal diffusion methods extend video generation beyond a fixed temporal window, including Diffusion Forcing~\citep{chen2024diffusion} and history-guided generation~\citep{song2025historyguided}. FramePack~\citep{zhang2025framepack} compresses previous frames into a bounded context, while MilliVid~\citep{chandratreya2026millivid} uses hierarchical latents and coarse-to-fine rollout to preserve long-range structure. These approaches reduce the cost of long-horizon generation, but recall stays indexed by recency: a region observed long ago is compressed or dropped regardless of whether the camera is about to return to it.\\
\noindent\textbf{Explicit spatial memory.} Several methods equip video generators with explicit  scene representations. Gen3C~\citep{ren2025gen3c}, Lyra and Lyra~2.0~\citep{bahmani2026lyra,shen2026lyra}, MosaicMem~\citep{mosaicmem2026}, and geometry-aware world models~\citep{wei2026gimworld,xiao2026worldmem} use reconstructed geometry, cached views, or spatial representations to retrieve information relevant to the current camera. Closely related ideas also appear in reconstruction systems with persistent or online state~\citep{cut3r,wang2025spann3r,chen2026ttt3r,zhuo2026streaming}. Such spatial memories naturally support revisits, but tie recall to an explicit reconstruction or cache whose representation and storage must be maintained as the explored scene grows. We instead retain spatially relevant information in a fixed-capacity learned state.\\
\noindent\textbf{Implicit and recurrent memory.}
State-space models~\citep{mamba,mamba2} compress arbitrarily long sequences into recurrent states at linear cost and have been adapted to long-context video~\citep{Po_2025_ICCV}. Gated-slot attention~\citep{gsa} and sparse-slot memories such as Raven~\citep{raven2026} partition the state into selectively updated slots, so information persists when unrelated inputs arrive. Learned states can also support geometric queries: CUT3R~\citep{cut3r} predicts pointmaps from its recurrent state at queried viewpoints. GIM-World~\citep{wei2026gimworld} compresses past observations into a fixed-size representation with camera-queryable geometry supervision. Yet, CUT3R requires explicit 3D supervision and GIM-World does not maintain a low-cost recurrent memory. \\
\method{} focuses on geometric addressing within the recurrence itself: camera pose governs autoregressive memory updates and readouts. At generation time, a target-camera query decodes a coarse full-frame draft to condition a chunk-causal video diffusion model only on the relevant information.
}

\section{Method}
\begin{wrapfigure}{r}{0.5\textwidth}
    \vspace{-20pt}
    \begin{center}
        \includegraphics[width=0.5\textwidth]{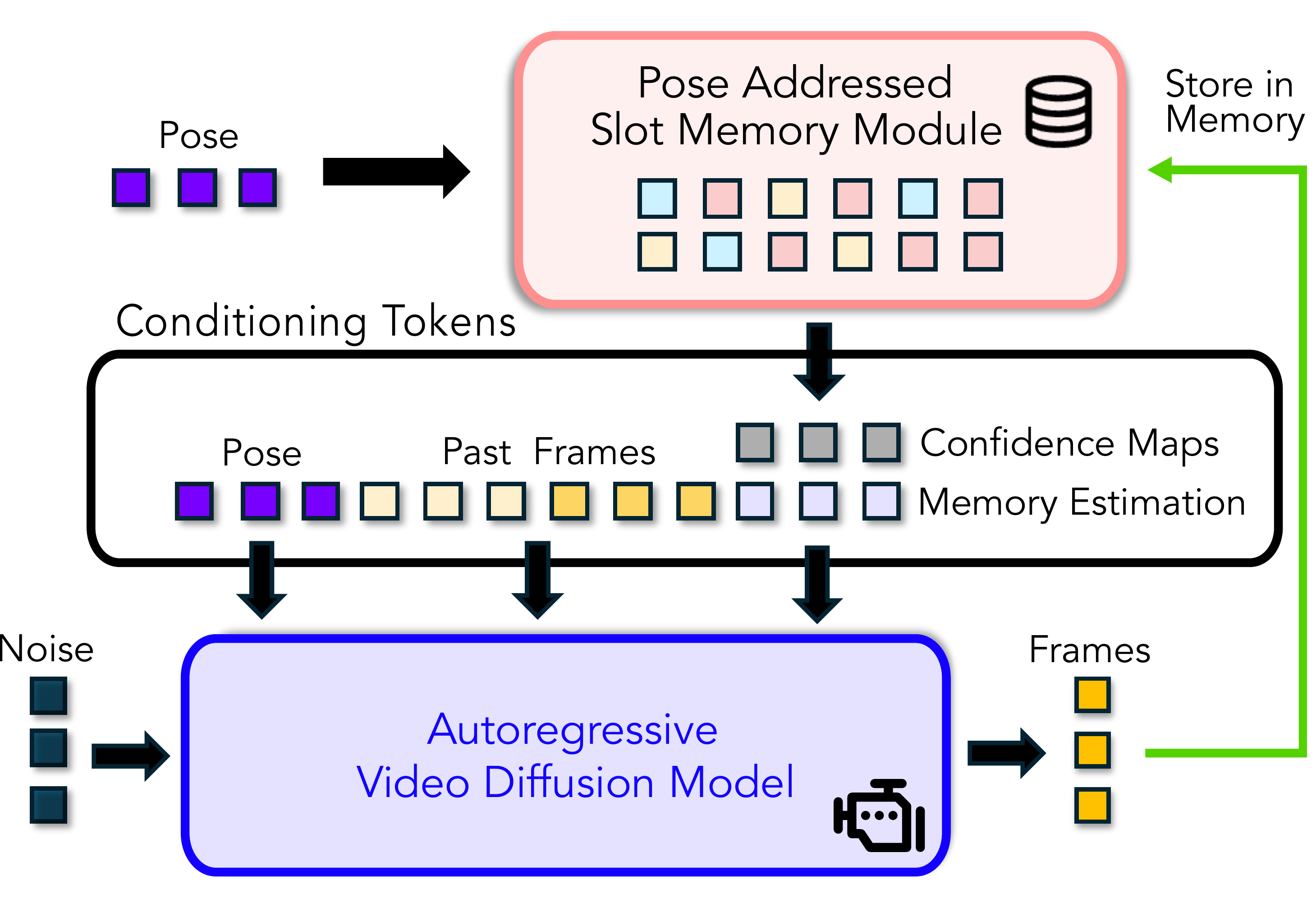}
    \end{center}
    \vspace{-8pt}
    \caption{\textbf{\method{} inference overview.}}
    \label{fig:overview}
    \vspace{-12pt}
\end{wrapfigure}
Our aim is the following: given a viewpoint trajectory through a scene, generate the video along it such that whenever a viewpoint re-observes a part of the scene, the generated content agrees with the history.\\
\noindent\textbf{Overview.} As shown in Fig.~\ref{fig:overview}, \method{} achieves this with two components. A recurrent slot memory stores content under pose-derived keys and, when queried with a target camera, returns a coarse draft of that view together with a confidence estimate in latent space.
A chunk-causal diffusion model sharpens the draft into video, trusting each cell in proportion to the memory's confidence. Generated chunks are written back into the memory, so rollouts of arbitrary lengths run at a constant memory size.
A detailed introduction to the individual components as well as the preliminaries follow below.

\subsection{Preliminaries}

\paragraph{State-space models.}
A state-space model processes a sequence through a recurrent state $\vh_t$ of fixed size, $\vh_t = \mA_t \vh_{t-1} + \mB_t \vu_t$ with output $\vy_t = \mC_t \vh_t$, so that arbitrarily long histories are compressed at $O(1)$ cost per token.
Structured SSMs \citep{s4} parameterize $(\mA, \mB, \mC)$ for stable long-range propagation, and selective variants \citep{mamba, mamba2} make them input-dependent, letting the model choose per token what to write and what to forget. The state is nonetheless a decaying summary ordered by time: what can be recovered from $\vh_t$ about an input $\vu_{t-n}$ diminishes with $n$. This is the property our memory design must overcome, and the slot-structured recurrence below is the halfway point between such dense states and full attention.

\paragraph{State-space models with Gated Slot Memory.}
Gated linear attention \citep{gla, mamba2} maintains a recurrent key--value state with a learned decay, and slot-structured variants~\citep{gsa} replace the single dense state with $M$ discrete slots per head, each holding a key--value pair updated by an exponential moving average
$\mH^{\{k/v\}}_m \leftarrow \exp(g)\, \mH^{\{k/v\}}_m + s\, \{\vk/\vv\}$
where $g \le 0$ is a per-head log-decay and $s$ a write gate. A read with query $\vq$ forms $\valpha = \softmax(\vq \cdot \mH^k)$ over the $M$ slots and returns $\vo = \valpha \cdot \mH^v$. Unlike a dense SSM state, a slot is an addressable unit: what a read returns is decided by the match between the query and the keys under which content was written, not by when the content arrived. We adopt this recurrence as the substrate of our memory, and describe our modifications in \cref{sec:memory}.

\paragraph{Latent Video Diffusion.}
Diffusion models \citep{ho2020denoising,song2021scorebased} generate by learning to invert a gradual noising of the data: a sample $\vx_0$ is mixed with noise $\veps \sim \gN(\vzero, \mI)$ to $\vx_\tau$ at level $\tau \in [0,1]$, and a network is trained to recover the clean sample, for instance by predicting the noise,
    $\Ls_{\mathrm{diff}} = \E_{\vx_0, \veps, \tau}\left[\, \lVert \veps_\vtheta(\vx_\tau, \tau, \gC) - \veps \rVert_2^2 \,\right],$
with $\gC$ the conditioning. Generation runs the reverse process from pure noise, and flow matching \citep{lipman2022flow} is an equivalent velocity parameterization of the same family. We operate in the latent space of a frozen video VAE \citep{kingma2014vae} that downsamples spatially and, depending on the autoencoder, temporally.

\subsection{Spatially-Structured Memory Design}

\label{sec:memory}

\begin{wrapfigure}{r}{0.5\textwidth}
    \vskip -0.15in
    \centering
    \includegraphics[width=0.5\textwidth]{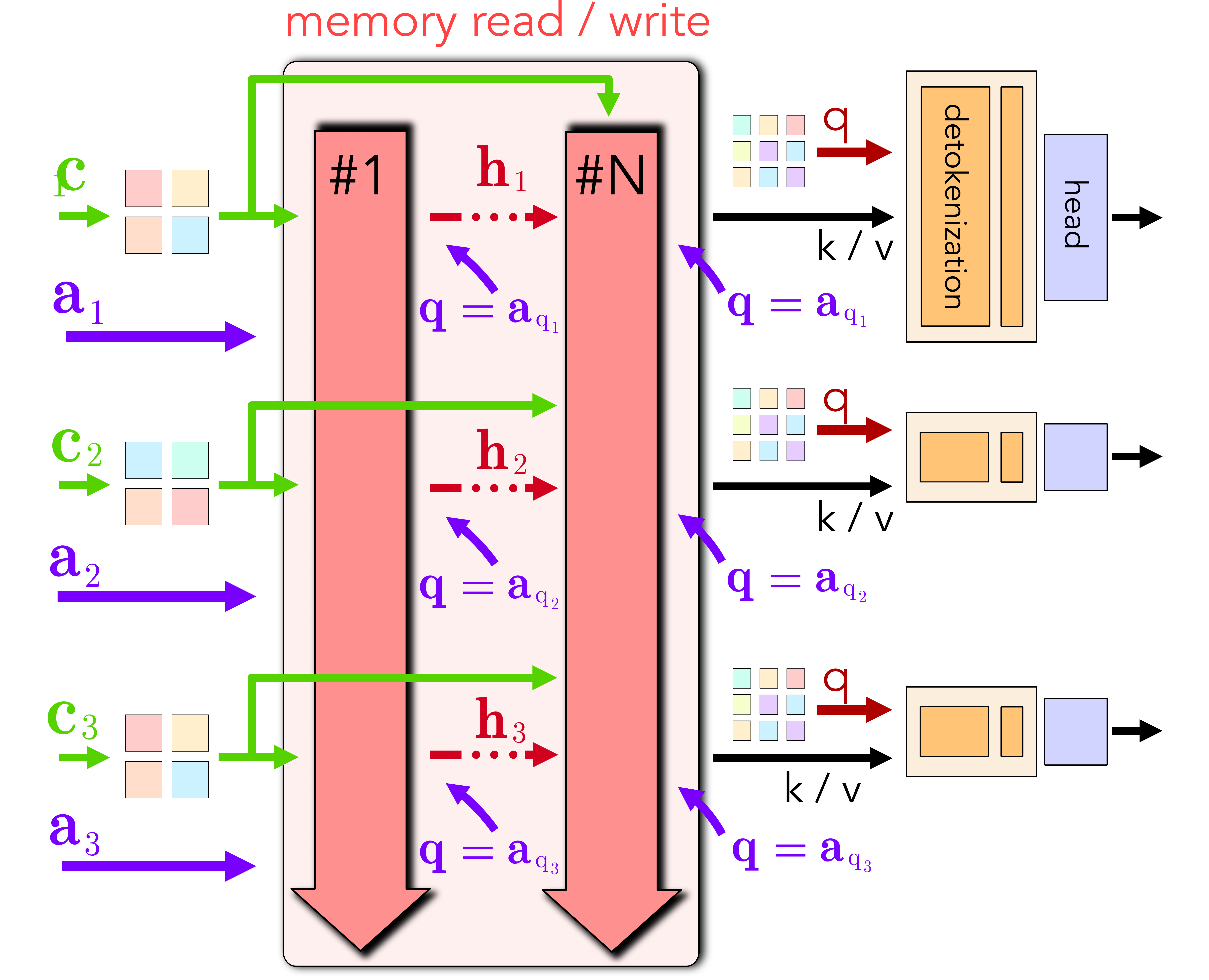}
    \caption{\textbf{\method{} memory module.} Write and read paths of the slot memory.}
    \label{fig:memory}
    \vskip -0.1in
\end{wrapfigure}

 The memory is a bank of $M$ slots per head across $N_h$ heads, written as posed latent frames arrive and read by presenting a target camera. Its design enforces one separation: everything that decides \emph{where} content goes (keys, queries, and routing) is computed from pose alone, while \emph{what} is stored, and how strongly it overwrites, is computed from content. Pose enters as data on the addressing side, as an absolute code matched between writer and reader, and as an operator on the content side, where the relative camera between writer and reader transports stored values and warps the decoder's attention. Fig.~\ref{fig:memory} summarizes the write and read paths. Details beyond this section are given in~\cref{app:impl}.

\vspace{-.5em}
\paragraph{Setup and tokenization.}
\begin{wrapfigure}{r}{0.3\textwidth}
    \centering
    \includegraphics[width=0.3\textwidth]{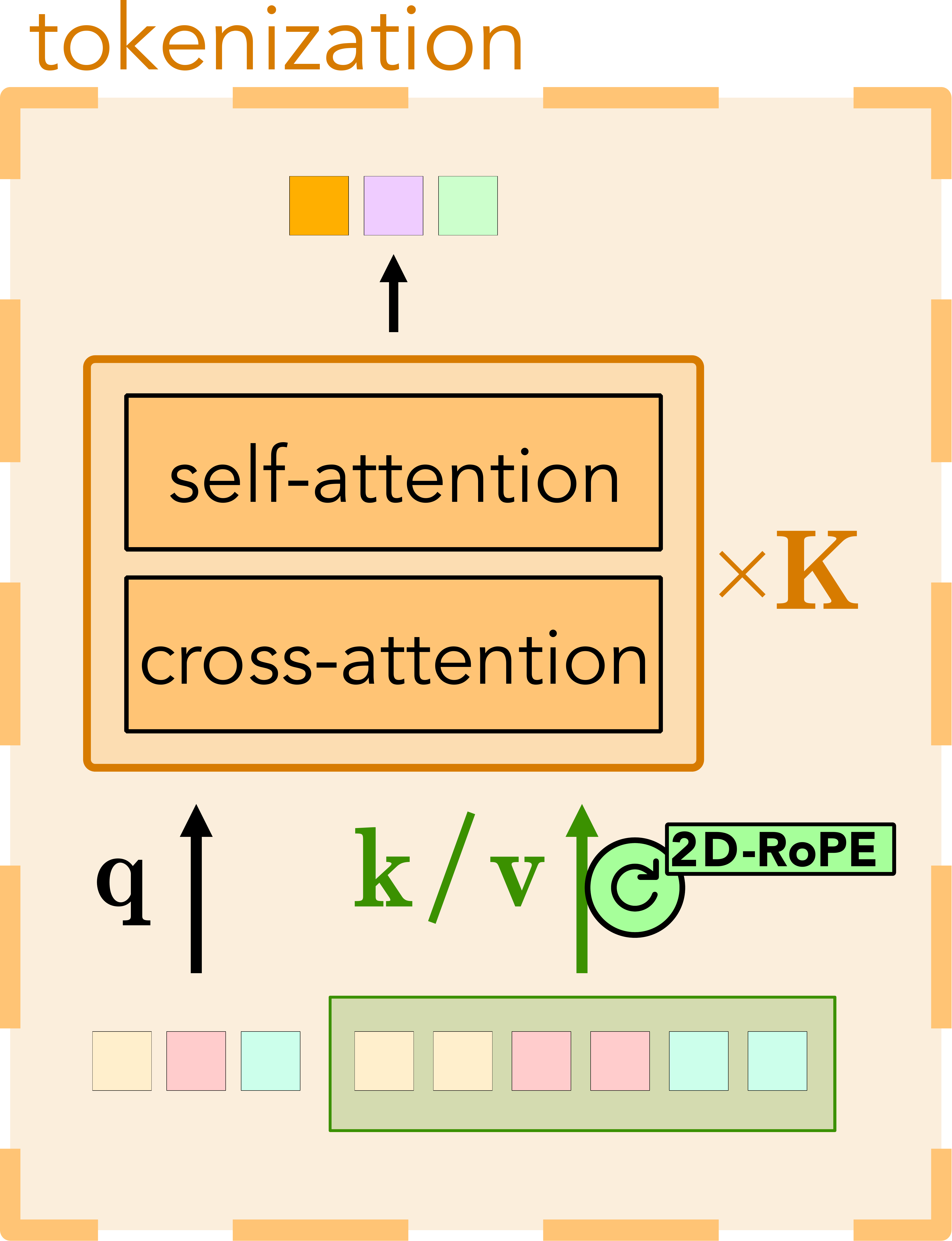}
    \caption{\textbf{Tiled tokenization.} Write token $j$ queries only the patches of tile $j$.}
    \label{fig:tokenization}
\end{wrapfigure}
A clip is $N$ posed video frames $\gX = \{\vx_i\}$ with cameras $\pi_i = (\mE_i, \mK_i)$, the extrinsics $\mE_i \in SE(3)$ rebased so that $\mE_0 = \mI$ and normalized in scale. A frozen video VAE encodes the clip to $T$ latent frames $\vz_t \in \R^{C \times H_z \times W_z}$, and everything downstream runs at the latent rate.
Each latent frame is patch-embedded into $P$ tokens and compressed by a tiled tokenizer (\cref{fig:tokenization}): the frame is divided into $J$ fixed image tiles, and write token $j$ cross-attends on the patches of tile $j$, so $\vc_{t,j}$ summarizes a fixed region of the frame before the per-head value projection. No spatial or temporal embedding is added to a write token, and temporal order is represented only by the recurrence itself. 

\paragraph{Pose addressing.}
The address of a region is computed based on viewing geometry alone, i.e., content
never enters it. For each cell of a coarse grid we form a footprint ray field given
by the Pl\"ucker coordinates of the ray,
\begin{equation}
    \vr_i = \Big[\,\vp_i \times \vd_{i,uv}\;;\;\vd_{i,uv}\,\Big]_{u,v} ,
    \label{eq:ray}
\end{equation}
with $\vp_i$ the camera center and $\vd_{i,uv}$ the unit ray through cell $(u,v)$.
Camera centers are expressed in units of a scene-scale normalizer $\sigma$, keeping
the moment term in a stable input range across scenes.
The raw geometric field is what lets displaced and co-observing queries land near the writes that cover them, and we verify this in \cref{sec:analysis-ablations}.
A convolutional encoder $\gE_{\mathrm{cam}}$ maps the field to an address grid $\mA_t \in \R^{D \times H_z \times W_z}$, and a second tiled tokenizer $\phi_{\mathrm{cam}}$ reduces it to one address per token and head,
\begin{equation}
    \left[\va_{h,t,1}, \dots, \va_{h,t,J}\right] = \phi_{\mathrm{cam}}(\mA_t),
    \qquad \va_{h,t,j} \in \R^{d_a},
    \label{eq:addr-pool}
\end{equation}
Thus, region token $\va_{h,t,j}$ carries address information for the same image tile as content token, $\vc_{t,j}$. Write keys and read queries are per-head identity-initialized linear functions of the address alone,
    $\vk^{h}_{t,j} = \mW^{h}_{k}\,\va^{h}_{t,j},
    \;
    \vq^{h}_{t^{\ast}\!,j} = \mW^{h}_{q}\,\va^{h}_{t^{\ast}\!,j},
    \label{eq:key-query}$
so the match between a query and a key depends only on the viewing geometry of the two regions, never on their content or on when they were written. 
When the VAE compresses time, each latent frame inherits $f$ cameras, and $\gE_{\mathrm{cam}}$ folds them causally at the VAE's own rate, so the address of latent frame $t$ depends on no later camera.

\paragraph{Routed writing.} \label{par:routed-writing}

\begin{wrapfigure}{r}{0.5\textwidth}
    \centering
    \includegraphics[width=0.5\textwidth]{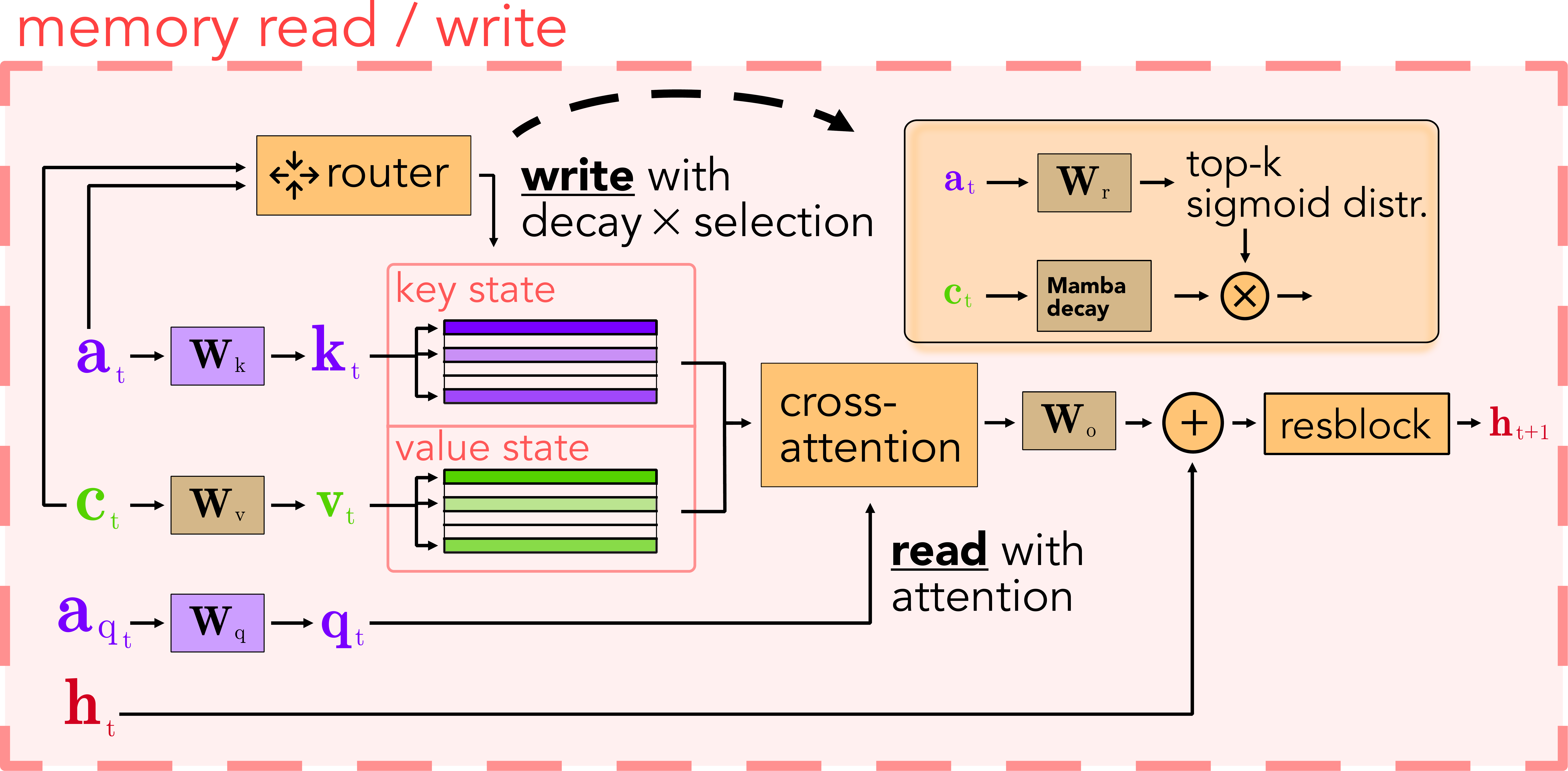}
    \caption{\textbf{Slot read and write.} Address routes the write; Read is a cross-attention over slot states.}
    \label{fig:dual-memory}
    \vskip -0.1in
\end{wrapfigure}

Every token write touches a single slot and leaves the rest unchanged. Each slot learns, through a fixed embedding, a region of pose space it is responsible for (the rows of a linear router $\mW_r$), and an incoming address activates the slot per head whose region matches it best, with strength $r_m \in \left[\,0,1\,\right)$ given by the match (\cref{fig:dual-memory}).
Scoring against fixed embeddings rather than against the slots' current contents keeps each slot stable over time and improves long-horizon recall, irrespective of what was written in between.
Before writing, the value is canonicalized with the inverse writing camera projection, $\tilde\vv$, so content observed from different viewpoints accumulates in a shared frame and can be re-posed to the target camera when read~\citep{Miyato2024GTA,li2025cameras}.
A selected slot then blends new content into what it holds with a content-derived decay $\delta \le 0$,
\begin{equation}
    g_m = \delta\, r_m, \qquad s_m = 1 - e^{g_m}, \qquad
    \mH^{k}_m \leftarrow e^{g_m} \mH^{k}_m + s_m\, \vk, \quad
    \mH^{v}_m \leftarrow e^{g_m} \mH^{v}_m + s_m\, \tilde\vv,
    \label{eq:routed-write}
\end{equation}
A strong match ($r_m\approx1$)  updates the memory decisively, while the unselected slots have $r_m = 0$, hence $e^{g_m} = 1$ and $s_m = 0$: no decay, no write, i.e. their state passes through the step \emph{identically}.
A dense recurrent state decays every entry at every step, whereas slot memories tie forgetting to writes rather than elapsed time~\citep{gsa,raven2026}; addressing those writes by pose alleviates content decay, and content fades only when observations land at the \emph{same} address.

\paragraph{Reading and decoding.}
Reading from the memory essentially asks ``what is visible from here?''. The target camera is passed through the same address encoder used for writing, so the question is posed in the language the answers were filed under: the query $\vq$ is scored against all slot keys and the memory returns a blend of slot contents, $\vo = \mathrm{softmax}(\beta\, \vq \cdot \mH^k)\cdot \mH^v$, with a learned per-head temperature $\beta$. 
The retrieved content is then rendered from the queried viewpoint: because values were canonicalized when written, the target camera re-poses the read into its own frame, effectively transporting content from the writing viewpoints to the query view. A cross-attention block lets each region of the target view pull what it needs from the readout, and a detokenizer expands the result to a coarse estimate and uncertainty~\citep{wang2024dust3r,cut3r} (details in \cref{app:impl}). 

\subsection{Memory-Conditioned Video Generation}
\label{sec:generation}

The generator's task is to turn the memory's coarse draft into video. It is a diffusion transformer that sees the memory through an interface composed of a coarse latent and its confidence mask. The current latent, memory prediction, and its confidence mask are channelwise concatenated before patch embedding similar to \cite{wan2025}, giving each token a spatially aligned memory estimate.

\paragraph{Chunk-causal attention.}
The model generates $N$ latent frames (a \textit{chunk}) at a time. A fixed mask governs all blocks: a \textsc{gen} token may attend to the $W$ clean context frames before its chunk, and its own chunk bidirectionally. 
Each denoising step is then a compact forward over the few frames a chunk can legally see rather than the full clip, so the cost of a step depends on $(W+N)$, never on how long the video has run. For a chunk starting at frame $s$, the visible sequence is
\begin{equation*}
    \underbrace{\;x_{s-W} \,\cdots\, x_{s-1}\;}_{\substack{\textsc{context} - \text{denoised}}}
    \;\big|\;
    \underbrace{\;\big(x^{\tau}_{s},\, \hat x_{s},\, 1{-}\tau_{c,s}\big) \,\cdots\, \big(x^{\tau}_{s+N-1},\, \hat x_{s+N-1},\, 1{-}\tau_{c,s+N-1}\big)\;}_{\substack{\textsc{gen} - \text{noised frame, draft, and trust as channels}}}
\end{equation*}
Further DiT, training and dataset-specific details are given in \cref{sec:implementation-details} and \cref{app:impl}.

\subsection{Training}
\label{sec:training}

We train in two stages, with a frozen domain-specific autoencoder throughout (\cref{tab:instantiation}). The memory is trained first, alone and as a plain regressor, so that its reads are verified retrieval before any generator depends on them. The diffusion transformer is then trained on top of the frozen memory.

\paragraph{Memory training.}
The memory is trained through a single interface: having written the observed frames $\gO_t$ into its state, it is handed one query pose and must return the latents observed there, $\hat z = f_\theta(z_{\gO_t}, \pi_{\gO_t};\, \pi_{\tau(t)}) \approx z_{\tau(t)}$. Everything the memory must learn is expressed by the choice of $\tau$ and $\gO$: revisits ($C_{\text{rev}}$), answerable only by recall, co-observed novel views ($C_{\text{nvs}}$), answerable only by fusing stored viewpoints, past ($C_{\text{past}}$) and current frames ($C_{\text{cur}}$), and held-out spans ($C_{\text{ho}}$), a quarter of each clip excluded from the write and rendered purely from surrounding observations. Recall and synthesis are the same read, differing only in where the target lies relative to what was written. Reading never modifies the state, so one written prefix can be queried at arbitrarily many cameras and all targets are supervised in parallel. The recurrence itself trains as a parallel scan, without unrolling.

\paragraph{Objective.}
The loss separates what the memory must produce from how it must be addressed,
\begin{equation}
    \gL \;=\; \gL^{\mathrm{conf}}_{\mathrm{read}} \;+\; \lambda_{\mathrm{wf}}\,\gL^{\mathrm{conf}}_{\mathrm{wf}} \;+\; \gL_{\mathrm{addr}},
    \hspace{0.5em} 
    \gL^{\mathrm{conf}} = \E\left[\,\vc_{\tau(t)}\,\big(\left(\vx_{\tau(t)} - \hat{\vx}_{\tau(t)}\right)^2 + \epsilon\big) - \alpha \log \vc_{\tau(t)}\,\right]
    \label{eq:mem-loss}
\end{equation}
where $\hat{\vx}_{\tau(t)}$ and $\vc_{\tau(t)}$ are the latent and the confidence the memory predicts for the queried pose, $\vx_{\tau(t)}$ the latent actually observed there.
The first two terms are the confidence-aware regression loss~\citep{wan2018confnet}, which lets the model lower its confidence where it cannot know the answer: $\gL^{\mathrm{conf}}_{\mathrm{read}}$ scores the causal read, and $\gL^{\mathrm{conf}}_{\mathrm{wf}}$ is a write-free pass over the finished clip's frozen state, exactly the read-only access the generator will later perform and never part of the causal rollout. 
The address term, $\gL_{\mathrm{addr}}$, keeps the code geometric without collapse: an anchor makes each token's viewing geometry recoverable from its address, and a deliberately weak VICReg spread~\citep{bardes2022vicreg} raises the rank of the address space, while keeping the neighbourhood structure. 
Term definitions and weights are fleshed out in \cref{app:ssm-training}.

\paragraph{Generator training.}
The transformer is trained with the flow-matching loss~\citep{lipman2022flow} on the generation rows only. Training prepares it for two realities of rollout. The history it conditions on will be self-generated and imperfect, so the clean context is sometimes noised, with the true level stamped through the modulation. Additionally, the draft it is handed will sometimes also be wrong, so drafts are deliberately corrupted, from single cells to the whole stream.
Training details and optimizer settings for both stages are listed in App.~\ref{app:impl}.

\section{Experiments}

\paragraph{Setup.}
We evaluate on \textit{Memory Maze}~\citep{pasukonis2022memorymaze,tunyasuvunakool2020}, procedurally generated egocentric 3D mazes at $64\times64$, and \textit{Minecraft}~\citep{yan2023temporally,guss2019minerl} open-terrain trajectories at $256\times256$ resolution (\cref{fig:topdowns}). Memory Maze is our primary testbed where revisits are enforced by construction, the maze layout gives an unambiguous ground truth for what should be recalled, and the low resolution makes ablations affordable. Minecraft tests whether the same mechanism holds at scale, on higher resolution action-conditioned generation.
For \textit{Minecraft}, we adopt 600-frame rollouts from the training dataset of~\cite{chandratreya2026millivid} and we generate 401-frame rollouts to train on \textit{Memory Maze} while we evaluate at longer horizons. 
Memory Maze retraces the observed trajectory in reverse, whereas Minecraft continues along the action-conditioned path. 
Following the protocol of~\cite{chandratreya2026millivid}, these paths are sampled to close loops over the context, so reconstruction metrics primarily measure recall of observed content.
The two instantiations share the interface of \cref{sec:generation} but differ in backbone, autoencoder, and resolution (\cref{tab:instantiation}; details in \cref{app:protocols}).\\
\paragraph{Metrics.} 
For reconstruction we report mean per-frame PSNR, SSIM \citep{wang2004ssim} with $11\times11$ window of $\sigma = 1.5$ and LPIPS~\citep{zhang2018perceptual} with AlexNet features. 
For generation, we report FID \citep{heusel2017fid} (InceptionV3, $2048$-D) and FVD \citep{unterthiner2018fvd} (I3D, $1024$-D).

\subsection{Implementation} \label{sec:implementation-details}
\paragraph{Memory design.}
We instantiate \method{} with a GSA slot recurrence~\citep{gsa} using Mamba-2 decay~\citep{mamba2} per slot, adapted from Raven~\citep{raven2026} for pose-addressed video memory. 
Full architectural and training details are given in \cref{app:impl,app:ssm-training}.\\
\paragraph{Generator design.}
We instantiate \method{} with two diffusion-transformer backbones to evaluate the memory across different generation settings: a camera-conditioned Wan2.1-style model~\citep{wan2025,li2025cameras} on \textit{Memory Maze} and an action-conditioned DiT-B~\citep{peebles2023scalable,chandratreya2026millivid} on \textit{Minecraft}. Both start from a chunk-causal generator without memory, trained by us over the full clean history on \textit{Memory Maze} and taken from MilliVid's released $7$-frame rollout checkpoint on \textit{Minecraft}, and are then finetuned with \method{}-SM. 
Given an observed prefix, the generator generates chunk by chunk along the control signal, and writes its own generations back into memory. 
Full architectural details are provided in \cref{app:protocols}.

\subsection{3D Consistent Video Generation}

\paragraph{Baselines.}
On \textit{Memory Maze} we compare against causal attention over the full history, its 100-frame windowed variant, DFoT~\citep{song2025historyguided}, and GeoCache, our reimplementation of the geometry-cached memory of Lyra~2.0~\citep{shen2026lyra} after training faithful implementations based on their official codebases. We additionally compare against our StateSpaceDiffuser~\citep{savov2026statespacediffuser} reimplementation, a capacity-matched action-conditioned recurrent memory, which degrades rapidly on longer trajectories (see \cref{app:additional-experiments}).
On \textit{Minecraft} we instead compare against released checkpoints throughout, namely MilliVid~\citep{chandratreya2026millivid}, the FramePack~\citep{zhang2025framepack} reimplementation distributed with it, and its 7-frame full-resolution rollout, each run through its authors' own sampling code.
Implementation details for all baselines are given in \cref{app:protocols}.

\paragraph{Generation results.}
Across both environments, \method{} maintains substantially stronger long-horizon reconstruction fidelity than temporal-context baselines while remaining markedly more compute-efficient than the strongest long-context alternatives (\cref{tab:video-generation,tab:video-generation-minecraft}).
On \textit{Memory Maze}, reconstruction quality improves up to the training horizon and degrades only gradually as the trajectory grows longer, whereas all temporal-context baselines deteriorate more.
Across the longer horizons, \method{} maintains a clear margin in reconstruction quality while requiring substantially less compute than DFoT and GeoCache.
On \textit{Minecraft}, \method{} achieves the highest PSNR and SSIM across all horizons and the best LPIPS in three of four bands, while trailing MilliVid on FID and on FVD at intermediate horizons, at roughly $2.7\times$ lower compute.
Qualitative examples are shown in \cref{fig:qualitative} for \textit{Minecraft} and \cref{fig:qualitative-memory-maze} for \textit{Memory Maze}.
\definecolor{qgrowgray}{RGB}{58,58,58}

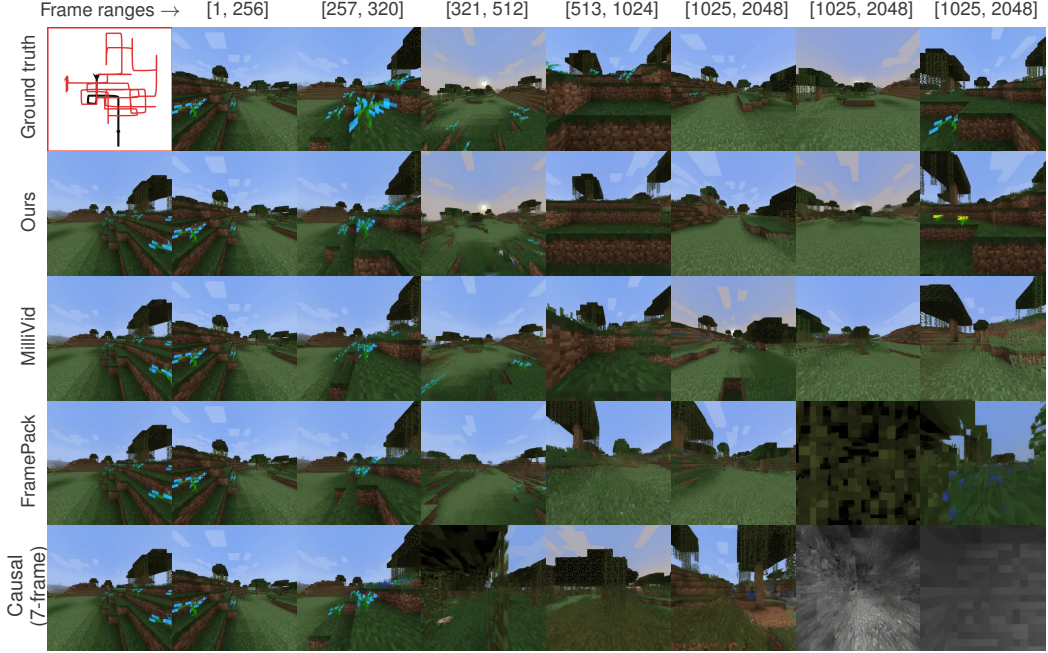
\begin{figure}[t]
  \centering

\newcommand{\qqqqqgImageRoot}{Figures/raw_pngs/experiments/104}

\newcommand{\qqqqqgPanel}[1]{%
  \includegraphics[
    width=1.65cm,
    height=1.65cm,
    keepaspectratio=false
  ]{#1}%
}

  \begin{tikzpicture}[x=1cm,y=1cm]
    \def\colw{1.65}
    \def\rowh{1.65}
    \def\leftpad{0}

    \foreach \range [count=\i from 1] in {
      {[1, 256]},
      {[257, 320]},
      {[321, 512]},
      {[513, 1024]},
      {[1025, 2048]},
      {[1025, 2048]},
      {[1025, 2048]}
    }
    {
      \node[
        font=\scriptsize\sffamily,
        qgrowgray,
        anchor=south,
        inner sep=0pt,
        xshift=0.05cm,
        yshift=0.06cm
      ] at ({\leftpad+(\i+0.5)*\colw},0.04) {\range};
    }

    \node[
      font=\scriptsize\sffamily,
      qgrowgray,
      anchor=south west,
      inner sep=0pt
    ] at (-0.1,0.08) {Frame ranges $\to$};

    \foreach \rowfolder [count=\r from 0] in {row1,row2,row3,row4,row5}{
      \ifcase\r
        \def\files{01_topdown.png,02_Image34.png,03_Image35.png,04_Image36.png,05_Image37.png,06_Image38.png,07_Image39.png,08_Image40.png}
      \or
        \def\files{01_Image25.png,02_Image26.png,03_Image27.png,04_Image28.png,05_Image29.png,06_Image30.png,07_Image31.png,08_Image32.png}
      \or
        \def\files{01_Image17.png,02_Image18.png,03_Image19.png,04_Image20.png,05_Image21.png,06_Image22.png,07_Image23.png,08_Image24.png}
      \or
        \def\files{01_Image9.png,02_Image10.png,03_Image11.png,04_Image12.png,05_Image13.png,06_Image14.png,07_Image15.png,08_Image16.png}
      \or
        \def\files{01_Image1.png,02_Image2.png,03_Image3.png,04_Image4.png,05_Image5.png,06_Image6.png,07_Image7.png,08_Image8.png}
      \fi

      \ifnum\r=0
        \node[
          rotate=90,
          anchor=center,
          font=\scriptsize\sffamily,
          qgrowgray
        ] at (-0.25,{-0.8-\r*\rowh}) {Ground truth};
      \else\ifnum\r=1
        \node[
          rotate=90,
          anchor=center,
          font=\scriptsize\sffamily,
          qgrowgray
        ] at (-0.25,{-0.8-\r*\rowh}) {Ours};
      \else\ifnum\r=2
        \node[
          rotate=90,
          anchor=center,
          font=\scriptsize\sffamily,
          qgrowgray
        ] at (-0.25,{-0.8-\r*\rowh}) {MilliVid};
      \else\ifnum\r=3
        \node[
          rotate=90,
          anchor=center,
          font=\scriptsize\sffamily,
          qgrowgray
        ] at (-0.25,{-0.8-\r*\rowh}) {FramePack};
      \else
        \node[
          rotate=90,
          anchor=center,
          font=\fontsize{7.5}{5}\selectfont\sffamily,
          qgrowgray
        ] at (-0.25,{-0.80-\r*\rowh})
          {\shortstack{Causal\\(7-frame)}};
      \fi\fi\fi\fi

      \foreach \file [count=\col from 1] in \files{
        \pgfmathtruncatemacro{\x}{\col-1}
        \pgfmathtruncatemacro{\y}{\r}

        \node[
          inner sep=0pt,
          anchor=north west
        ] at ({\leftpad+\x*\colw},{-\y*\rowh})
        {\qqqqqgPanel{\qqqqqgImageRoot/\rowfolder/\file}};
      }
    }
  \end{tikzpicture}

  \caption{\textbf{Qualitative comparison on \textit{Minecraft}.} Generated frames
    for each method in \cref{tab:video-generation-minecraft}, on the same
    trajectory, at the 2048 frame horizon with 256 initial context frames.}
  \label{fig:qualitative}
\end{figure}

\begin{table*}[t]
\caption{\textbf{3D consistent video generation on \textit{Memory Maze}.} Methods generate the reverse leg from the observed forward trajectory in parentheses. Column groups are total trajectory length. FID pools a band's frames, FVD its $16$-frame clips.}
\vspace{-2mm}
\label{tab:video-generation}

\centering
\begin{scriptsize}
\begin{sc}
\setlength{\tabcolsep}{2.0pt}
\renewcommand{\metric}[1]{{\scriptsize #1}}

\begin{tabular}{@{}lcccccc@{\hspace{7pt}}cccccc@{}}
\multirow{2}{*}{Method}
&
\multicolumn{6}{c}{\normalfont\itshape $(201 + 21) + 201 = 423$ frames}
&
\multicolumn{6}{c}{\normalfont\itshape $(401 + 21) + 401 = 823$ frames}
\\

\cmidrule(lr){2-7}
\cmidrule(lr){8-13}

&
\metric{PFLOPs$\downarrow$}
& \metric{PSNR$\uparrow$}
& \metric{SSIM$\uparrow$}
& \metric{LPIPS$\downarrow$}
& \metric{FID$\downarrow$}
& \metric{FVD$\downarrow$}
&
\metric{PFLOPs$\downarrow$}
& \metric{PSNR$\uparrow$}
& \metric{SSIM$\uparrow$}
& \metric{LPIPS$\downarrow$}
& \metric{FID$\downarrow$}
& \metric{FVD$\downarrow$}
\\

\midrule

{\scriptsize Causal (100F)}
& \textbf{0.096}
& 25.0
& 0.809
& 0.131
& 6.39
& 4.06
& \textbf{0.193}
& 21.8
& 0.739
& 0.198
& 8.85
& 7.19
\\

{\scriptsize Causal (Full)}
& 0.117
& 30.0
& 0.891
& \underline{0.034}
& \underline{5.36}
& \underline{2.13}
& 0.299
& 24.3
& 0.789
& 0.100
& 7.96
& 4.61
\\

{\scriptsize GeoCache (GT-D)}
& 0.454
& 24.6
& 0.811
& 0.098
& 6.25
& 3.57
& 0.936
& 24.6
& 0.810
& 0.095
& 6.03
& 2.52
\\

{\scriptsize DFoT}
& 0.566
& \textbf{32.0}
& \textbf{0.919}
& \textbf{0.030}
& 5.74
& \textbf{1.89}
& 1.130
& \underline{28.4}
& \underline{0.865}
& \underline{0.079}
& \underline{5.54}
& \underline{1.89}
\\

\cmidrule{1-13}

{\scriptsize \method{}}
& \underline{0.108}
& \underline{31.1}
& \textbf{0.919}
& 0.040
& \textbf{5.22}
& 2.25
& \underline{0.213}
& \textbf{29.9}
& \textbf{0.900}
& \textbf{0.046}
& \textbf{5.08}
& \textbf{1.83}
\\

\bottomrule
\end{tabular}

\vspace{5pt}

\begin{tabular}{@{}lcccccc@{\hspace{7pt}}cccccc@{}}

\multirow{2}{*}{Method}
&
\multicolumn{6}{c}{\normalfont\itshape $(801 + 21) + 801 = 1623$ frames}
&
\multicolumn{6}{c}{\normalfont\itshape $(1601 + 21) + 1601 = 3223$ frames}
\\

\cmidrule(lr){2-7}
\cmidrule(lr){8-13}

&
\metric{PFLOPs$\downarrow$}
& \metric{PSNR$\uparrow$}
& \metric{SSIM$\uparrow$}
& \metric{LPIPS$\downarrow$}
& \metric{FID$\downarrow$}
& \metric{FVD$\downarrow$}
&
\metric{PFLOPs$\downarrow$}
& \metric{PSNR$\uparrow$}
& \metric{SSIM$\uparrow$}
& \metric{LPIPS$\downarrow$}
& \metric{FID$\downarrow$}
& \metric{FVD$\downarrow$}
\\

\midrule

{\scriptsize Causal (100F)}
& \textbf{0.394}
& 20.4
& 0.709
& 0.227
& 10.3
& 8.88
& \textbf{0.824}
& 19.1
& 0.677
& 0.266
& 11.6
& 11.1
\\

{\scriptsize Causal (Full)}
& 0.850
& 20.8
& 0.718
& 0.210
& 47.4
& 32.6
& 2.727
& 18.3
& 0.669
& 0.351
& 158.5
& 199.6
\\

{\scriptsize GeoCache (GT-D)}
& 1.872
& 25.3
& 0.822
& \underline{0.080}
& 5.71
& 2.04
& 3.772
& \underline{25.7}
& \underline{0.827}
& \underline{0.073}
& \underline{5.67}
& \underline{1.76}
\\

{\scriptsize DFoT}
& 2.259
& \underline{26.6}
& \underline{0.838}
& 0.103
& \underline{5.48}
& \textbf{1.60}
& 4.518
& 24.0
& 0.795
& 0.147
& 5.72
& 1.80
\\

\cmidrule{1-13}

{\scriptsize \method{}}
& \underline{0.430}
& \textbf{29.3}
& \textbf{0.890}
& \textbf{0.051}
& \textbf{5.03}
& \underline{1.66}
& \underline{0.900}
& \textbf{27.6}
& \textbf{0.863}
& \textbf{0.070}
& \textbf{5.00}
& \textbf{1.76}
\\

\bottomrule
\end{tabular}

\end{sc}
\end{scriptsize}
\end{table*}

\begin{table*}[t]
\caption{\textbf{3D consistent video generation on \textit{Minecraft}.} Methods generate frames $257$--$2048$ from $256$ ground-truth context frames. Column groups are distance past the context. FID pools a band's frames, FVD its $16$-frame clips.}
\label{tab:video-generation-minecraft}
\vspace{-2mm}

\centering
\begin{scriptsize}
\begin{sc}
\setlength{\tabcolsep}{2.2pt}
\renewcommand{\metric}[1]{{\scriptsize #1}}

\begin{tabular}{@{}lcccccc@{\hspace{7pt}}cccccc@{}}

\multirow{2}{*}{Method}
&
\multicolumn{6}{c}{\normalfont\itshape 1--64 frames}
&
\multicolumn{6}{c}{\normalfont\itshape 65--256 frames}
\\

\cmidrule(lr){2-7}
\cmidrule(lr){8-13}

&
\metric{PFLOPs$\downarrow$}
& \metric{PSNR$\uparrow$}
& \metric{SSIM$\uparrow$}
& \metric{LPIPS$\downarrow$}
& \metric{FID$\downarrow$}
& \metric{FVD$\downarrow$}
&
\metric{PFLOPs$\downarrow$}
& \metric{PSNR$\uparrow$}
& \metric{SSIM$\uparrow$}
& \metric{LPIPS$\downarrow$}
& \metric{FID$\downarrow$}
& \metric{FVD$\downarrow$}
\\

\midrule

{\scriptsize Causal (7F)}
& \textbf{0.72}
& 17.18
& 0.583
& 0.325
& \textbf{43.36}
& 64.37
& \textbf{2.62}
& 11.67
& 0.437
& 0.551
& 47.78
& 182.03
\\

{\scriptsize FramePack}
& 1.71
& 19.13
& 0.612
& 0.260
& 44.66
& 56.67
& 6.62
& 14.04
& 0.498
& 0.448
& 48.67
& 112.84
\\

{\scriptsize MilliVid}
& 2.40
& \underline{21.16}
& \underline{0.615}
& \underline{0.196}
& 44.68
& \underline{47.83}
& 8.95
& \underline{18.43}
& \underline{0.562}
& \textbf{0.284}
& \textbf{43.34}
& \textbf{44.28}
\\

\cmidrule{1-13}

{\scriptsize \method{}}
& \underline{0.91}
& \textbf{21.96}
& \textbf{0.647}
& \textbf{0.187}
& \underline{44.60}
& \textbf{45.86}
& \underline{3.46}
& \textbf{18.54}
& \textbf{0.571}
& \underline{0.294}
& \underline{44.01}
& \underline{56.00}
\\

\bottomrule
\end{tabular}

\vspace{5pt}

\begin{tabular}{@{}lcccccc@{\hspace{7pt}}cccccc@{}}

\multirow{2}{*}{Method}
&
\multicolumn{6}{c}{\normalfont\itshape 257--768 frames}
&
\multicolumn{6}{c}{\normalfont\itshape 769--1792 frames}
\\

\cmidrule(lr){2-7}
\cmidrule(lr){8-13}

&
\metric{PFLOPs$\downarrow$}
& \metric{PSNR$\uparrow$}
& \metric{SSIM$\uparrow$}
& \metric{LPIPS$\downarrow$}
& \metric{FID$\downarrow$}
& \metric{FVD$\downarrow$}
&
\metric{PFLOPs$\downarrow$}
& \metric{PSNR$\uparrow$}
& \metric{SSIM$\uparrow$}
& \metric{LPIPS$\downarrow$}
& \metric{FID$\downarrow$}
& \metric{FVD$\downarrow$}
\\

\midrule

{\scriptsize Causal (7F)}
& \textbf{7.77}
& 10.99
& 0.412
& 0.627
& 70.19
& 507.07
& \textbf{18.1}
& 10.51
& 0.392
& 0.700
& 109.60
& 963.68
\\

{\scriptsize FramePack}
& 19.7
& 11.97
& 0.444
& 0.540
& 58.12
& 228.78
& 45.9
& 11.21
& 0.411
& 0.601
& 64.31
& 333.38
\\

{\scriptsize MilliVid}
& 26.2
& \underline{16.05}
& \underline{0.522}
& \underline{0.371}
& \textbf{41.19}
& \textbf{56.37}
& 65.5
& \underline{13.84}
& \underline{0.468}
& \underline{0.477}
& \textbf{38.85}
& \underline{104.89}
\\

\cmidrule{1-13}

{\scriptsize \method{}}
& \underline{10.2}
& \textbf{16.38}
& \textbf{0.537}
& \textbf{0.370}
& \underline{44.55}
& \underline{72.61}
& \underline{23.8}
& \textbf{14.85}
& \textbf{0.503}
& \textbf{0.446}
& \underline{43.33}
& \textbf{101.75}
\\

\bottomrule
\end{tabular}

\end{sc}
\end{scriptsize}
\vspace{-1mm}
\end{table*}

\section{Analysis \& Ablations} \label{sec:analysis-ablations}

\begin{figure}[t]
    \begin{center}
        \centerline{\includegraphics[width=\columnwidth]{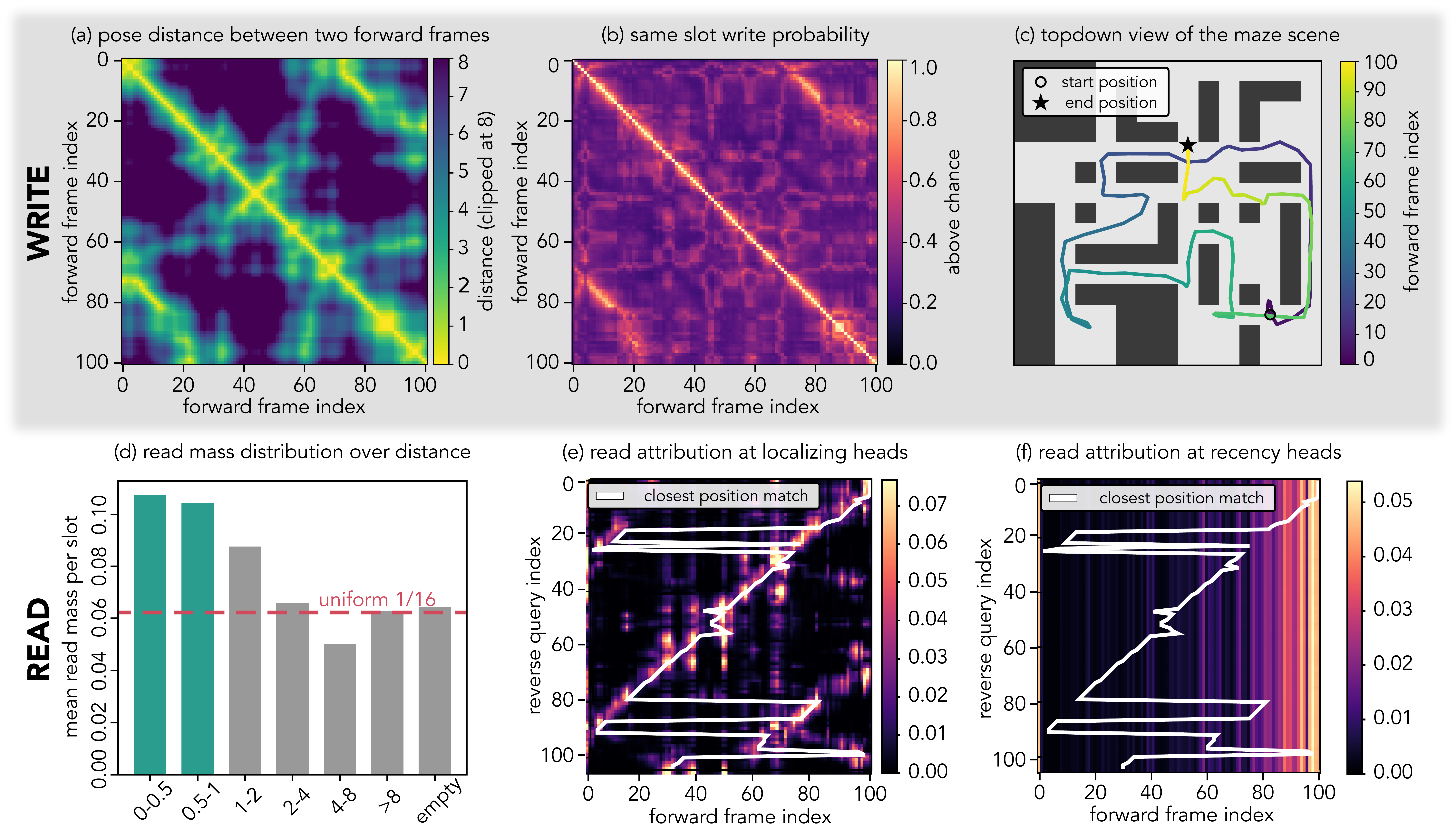}}
        \vspace{-.5em}
        \caption{\textbf{
        What the slot memory stores, and how a read recovers it.} \method{}-SM on $401$-frame maze episodes: the forward leg is written, the reverse leg read from the state. 
        \textbf{Write.} (a) Euclidean distance between two forward frame positions and (b) the probability that their region tokens land in the same slot.
        \textbf{Read.} (d) Mean read mass per slot against the distance from the query to the content that slot currently holds where content position is the decayed average of all content written to respective slot.
        (e,\,f) Attribution of each reverse read to the forward frames in memory slots where the white trace is the pose-nearest forward frame.}
        \label{fig:pose-addressing}
    \end{center}
    \vspace{-1em}
\end{figure}

\paragraph{Slot memory addressing by region tokens.}
Our slot memory addressing mechanism via region tokens allows for the decoupling of elapsed time and forgetting by writing only during revisits. In \cref{fig:pose-addressing}, we show that our design creates a strong overlap between the camera pose of a frame (a) and the memory slot to which it is being written (b). As a result, a slot only decays when re-routed. Reading the memory is a softmax-attention over the mixture of camera pose encodings, yielding a convex combination of write tokens, meaning that each query is attributable to the source frames. Accordingly, \cref{fig:pose-addressing} shows that slots holding the query's place draw $1.7\times$ uniform mass (c). Finally, we observe that not all heads behave similarly: In \cref{fig:pose-addressing}, we observe that while some heads are localizing heads for retrieving local information (e), others preserve a global attention pattern (f).\\
\paragraph{Ablations.} Table~\ref{tab:pose-enc-canonical} varies one component of \method{}'s memory module at a time, decoding frames from memory alone at queried poses. Due to perception-distortion trade-off~\citep{blau2018perception} and lack of generative stochasticity, the memory module alone can achieve high reconstruction scores at worse perceptual quality.  Replacing the routed slots with a dense state of the same size matches \method{} at training length
but degrades sharply beyond it, whereas \method{} holds its quality out to eight times that horizon: sparse writes allow extending memory horizon as a slot without further writes is left intact. Removing the SE(3) value transform costs quality at every horizon, indicating that stored content must be expressed relative to the querying camera to be read back from a new viewpoint. As expected, halving the memory heads reduces the retrieval capability in longer sequences.
\begin{table}[t]
    \caption{\textbf{Memory Only ablation results.} Reconstruction quality on \textit{Memory Maze}, changing one component of \method{} at a time. 
    \method{} applies the SE(3) value transform to the $64$ value channels of the head. All models are trained on 401 frame episodes.}
    \label{tab:pose-enc-canonical}
    \vspace{-2mm}
    \begin{center}
        \begin{scriptsize}
        \begin{sc}
          \setlength{\tabcolsep}{4pt}
          \begin{tabular}{@{}lccccccccc@{}}
            \multirow{2}{*}{Variant} & \multicolumn{3}{c}{\normalfont\itshape 401 frames} & \multicolumn{3}{c}{\normalfont\itshape 1601 frames} & \multicolumn{3}{c}{\normalfont\itshape 3201 frames} \\
            \cmidrule(lr){2-4} \cmidrule(lr){5-7} \cmidrule(lr){8-10}
            & \metric{PSNR$\uparrow$} & \metric{SSIM$\uparrow$} & \metric{LPIPS$\downarrow$} & \metric{PSNR$\uparrow$} & \metric{SSIM$\uparrow$} & \metric{LPIPS$\downarrow$} & \metric{PSNR$\uparrow$} & \metric{SSIM$\uparrow$} & \metric{LPIPS$\downarrow$} \\
            \midrule
            \method{}-Slot Memory & 32.7 & \winning{0.940} & 0.024 & \winning{32.4} & \winning{0.936} & \winning{0.024} & \winning{31.9} & \winning{0.931} & \winning{0.026} \\
            \cmidrule(lr){1-10}
            No SE(3) Value Warping & 31.5 & 0.924 & 0.031 & 30.0 & 0.906 & 0.041 & 29.6 & 0.900 & 0.045 \\
            \method{}-Dense Memory & \winning{32.9} & 0.939 & \textbf{0.020} & 26.5 & 0.852 & 0.074 & 21.5 & 0.756 & 0.191 \\
            Scaling Down Heads $(/2)$ & 32.2 & 0.935 & 0.028 & 24.5 & 0.810 & 0.146 & 22.9 & 0.772 & 0.192 \\
            \bottomrule
          \end{tabular}%
        \end{sc}
        \end{scriptsize}
        \vspace{-4mm}
    \end{center}
\end{table}

\section{Conclusions}
\vspace{-2mm}
In this work we presented SMART, a pose-addressed slot memory for long-horizon video generation that combines constant-size recurrent storage with spatially indexed recall. By separating pose-derived addressing from content storage, updating only the slots of the observed region, and canonicalizing stored values across viewpoints, SMART makes retention depend on where content was seen rather than when.
The retrieved view serves as a coarse draft that a chunk-causal diffusion model refines into video, drawing on stored scene content where it is available and on its generative prior elsewhere.
Across \textit{Memory Maze} and \textit{Minecraft}, this preserves scene-specific consistency over substantially longer horizons than temporal-context baselines while avoiding the growing storage and reconstruction dependence of explicit geometric memories. \\
\paragraph{Limitations \& Future Work.}
A central limitation of the current design is that fixed memory capacity bounds the spatial extent a model can retain. Although selective routing prevents information from decaying with elapsed time, sufficiently large environments eventually force different regions to reuse the same slots, causing previously stored content to be overwritten and forgotten. A natural direction for future work is therefore to make the memory capacity adaptive at inference time, for example, by allocating additional slots or memory banks online when the existing state becomes saturated, while preserving efficient pose-based addressing and retrieval. In addition, our current experiments use relatively modest generator backbones, scaling the approach to substantially larger video models is an important next step to assess how the memory mechanism interacts with stronger generative priors and higher visual fidelity.

\subsection*{AI use statement}
We used generative AI tools to assist with implementing and debugging research code, provide feedback on experimental methodology, identify relevant literature and revise the manuscript. The authors reviewed and revised the AI-assisted text, checked references against the original sources and reviewed and tested AI-assisted code. We take responsibility for the final manuscript, methods, results, and accompanying code, including all content produced with generative AI assistance.

\bibliography{arxiv_main}
\bibliographystyle{arxiv_main}
\newpage
\appendix
\section{Appendix}
\subsection{Relation to Linear Recurrent Models}
Closely related to our work, we draw similarities and differences with Raven~\citep{raven2026}, a gated-slot attention based linear-time sequence model.
We build on its state architecture where each head keeps a fixed number of slots, updated by the gated slot attention recurrence~\citep{gsa} with a learned decay, and a router assigns every token to the slots it writes, so the state stays constant in size however long the sequence grows.
Our implementation leverages the same underlying codebase from \texttt{fla}~\cite{yang2024fla} (\cref{app:impl}).
The difference lies in what the memory is asked to do.
In recurrent sequence models more broadly, queries, keys, values and routing are all projections of the current token's hidden state, and the state is read once, to predict the next token.

A camera-conditioned generator instead has to read the scene from any viewpoint, past or future, independently of the last frame written.
We therefore derive keys, queries and routing from the camera pose alone and values from the frame content to facilitate learning of pose-aligned memory slots rather than content similarity-based associations.
Each token writes a single slot, which remains stable until that place is observed again, and values are stored in a camera-independent frame to be re-posed by the target camera on read (\cref{par:value-transport}).
Finally, the output is not a next-token prediction but a draft of a target view, read from a state while the generator denoises and updates the state once the chunk is complete.

\subsection{Datasets} \label{app:datasets}

\begin{figure}[t]
    \begin{center}
        \centerline{\includegraphics[width=0.9\columnwidth]{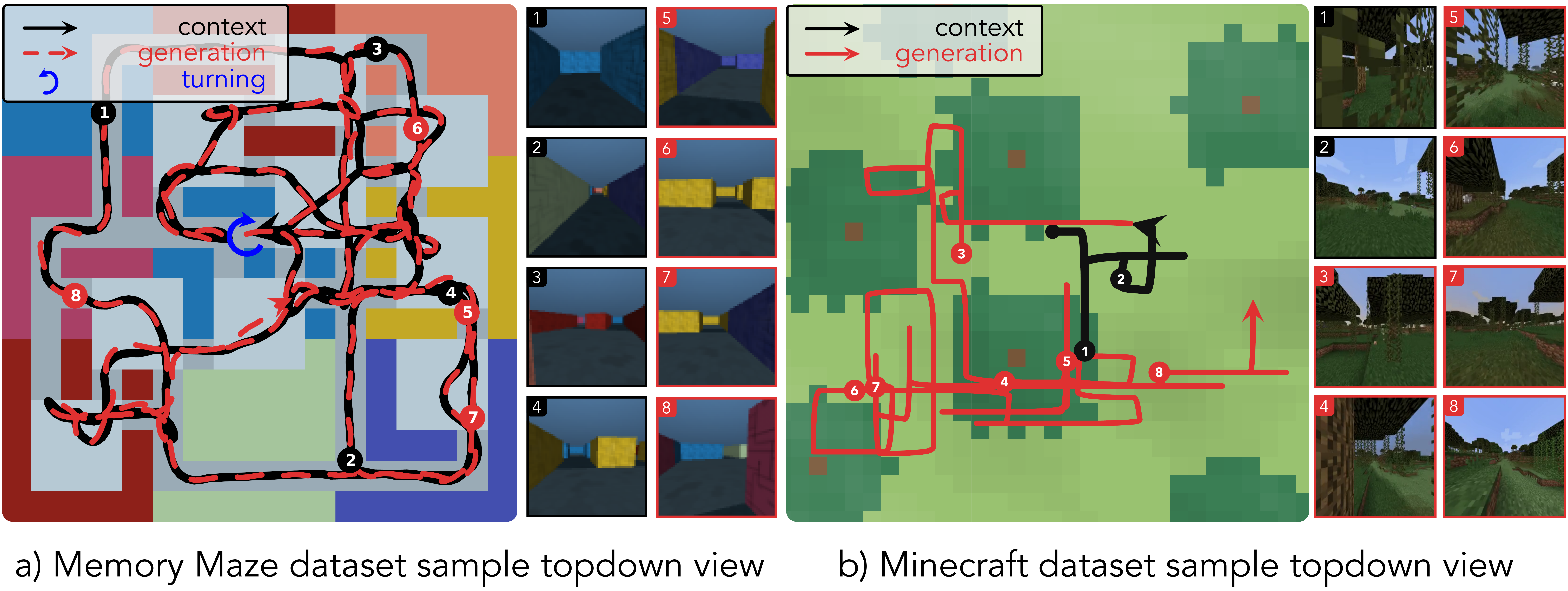}}
        \caption{\textbf{Evaluation trajectories.} Top-down view of one episode per dataset: 
        (a) a \textit{Memory Maze} forward--turn--return rollout, and (b) a \textit{Minecraft} continuation rollout.}
        \label{fig:topdowns}
    \end{center}
    \vspace{-2em}
\end{figure}

\paragraph{Memory Maze.}
Measuring retrieval needs trajectories that return to what they have already seen, which the released \textit{Memory Maze} episodes~\citep{pasukonis2022memorymaze} cannot guarantee: they record an RL agent exploring freely. 
We therefore roll out our own evaluation episodes in the same $15{\times}15$ environment on MuJoCo~\citep{todorov2012mujoco} and \texttt{dm\_control}~\citep{tunyasuvunakool2020}, with an imposed three-leg structure of a forward leg of $N$ frames, a $21$-frame in-place turn, and a return leg of $N$ frames, so an episode is $2N{+}21$ frames long (\cref{fig:topdowns}a).
The forward leg is a stochastic breadth-first coverage sweep, repeatedly heading for the nearest cell the route has not yet entered rather than following a shortest path, 
and the return leg then retraces the positions actually walked.
Frames are $64\times64$ RGB from a rigidly mounted first-person camera.

Each frame stores the agent position and heading. 
Camera extrinsics and intrinsics are derived from the pose at load time.
For training we re-render the released Memory Maze episodes~\citep{pasukonis2022memorymaze} through the same camera ($29{,}000$ train and $1{,}000$ evaluation), and cut into overlapping $401$-frame snippets so that the environment's goal spheres and target rings are hidden.

\paragraph{Minecraft.}
MilliVid's \textit{LoopCraft}~\citep{chandratreya2026millivid} provides our training data and the evaluation protocol we adopt, but its released test split is $1{,}024$ frames long and our claim concerns horizons beyond that.
We therefore generate a $2{,}048$-frame split with their released generator to collect MineRL~\citep{guss2019minerl} trajectories through the pipeline of TECO~\citep{yan2023temporally} where the agent moves forward or turns left/right, rendered at $512\times512$ and downsampled to $256\times256$.
Unlike our Memory Maze episodes these rollouts are continuations with no turn and no return leg, the model being given $256$ context frames and generating the remaining $1{,}792$ (\cref{fig:topdowns}b).
We keep~\citet{chandratreya2026millivid}'s loop-closure bias, drawing $300$ candidate action sequences per episode, scoring each by the Chamfer distance between the planned positions of the first $256$ frames and the remainder, and replaying one sampled uniformly from the ten scoring lowest.

MilliVid's \textit{LoopCraft} stores each episode's video, a one-hot action per frame, and the agent position and heading read directly from the simulator. To achieve pose-conditioning, we derive the camera poses based on the specifications of~\citet{guss2019minerl} and agent's pose.
For training we use $601$-frame rollouts, which is sufficient for the long-horizon extrapolation we evaluate.

\subsection{SMART Implementation Details}
\label{app:impl}

\paragraph{Tokenization.}
Each latent frame is cut into $P = H_z \cdot W_z / p^2$ patches of size $p$ and embedded by a strided convolution,
\begin{equation}
    \ve_{t,i} = \mathrm{Emb}(\vz_t)_i \in \R^{D}, \qquad i = 1, \dots, P,
    \label{eq:patch}
\end{equation}
which at $p{=}1$ is one token per latent cell. A tokenization module then compresses the frame's patch grid into $J \ll P$ \emph{write tokens}: $J$ learned queries attend the $P$ patch tokens, whose keys carry a two-dimensional rotary embedding over the $(H_z, W_z)$ grid~\citep{heo2024rotary}, and a causal self-attention over the $J$ queries makes every prefix $\vc_{t,1:j}$ a self-contained code of the frame \citep{asim2026scenetok},
\begin{equation}
    \gC_t=\left[\vc_{t,1}, \dots, \vc_{t,J}\right] = \mathrm{\phi_{content}}\!\left(\ve_{t,1}, \dots, \ve_{t,P}\right),
    \qquad \vc_{t,j} \in \R^{D}.
    \label{eq:pool}
\end{equation}
Write tokens are laid out frame-major into one sequence of length $S = TJ$, and the memory values are a linear projection of them, split across the $N_h$ heads,
\begin{equation}
    \vv_{h,t,j} = \left(\mW_v\, \vc_{t,j}\right)_h \in \R^{d}.
    \label{eq:value}
\end{equation}
$\vc_{t,j}$ carries neither a 3D-aware nor a temporal tag of its own via positional encodings so that the temporal order is represented only by the recurrence itself.

\paragraph{Memory implementation.} Our recurrent memory implementation uses the chunked-update GSA kernels of the \texttt{fla} package~\citep{yang2024fla}, inspired from Raven~\citep{raven2026}. 
In contrast to Raven, which derives keys, queries, router logits and values from the same layer input for language modeling, our design separates the pose-derived addressing pathway (keys, queries, router) from the content-derived value pathway, and writes $16$ tokens per frame rather than one per position based on the tokenization approach we took in~\cref{sec:memory}. 
Routing is top-$1$ with the raw sigmoid score of the winning slot as write amplitude, so the write gate is not renormalized across slots. 
Read-only operations are implemented by running the recurrence with zero keys, values and gates, which leaves the state unchanged.

\paragraph{Value transport.} \label{par:value-transport}
A view depends on where it was seen from, but a memory of the scene should not: values are therefore written in a shared camera-independent frame, the writing camera's inverse acting on their channels as an SE(3) transform, and re-posed by the target camera at read time~\citep{Miyato2024GTA,li2025cameras}. Because the slot read is linear in the values, a value written at $\pi_\tau$ and read at $\pi^\ast$ undergoes exactly the relative transform $\rho(\pi^\ast)\rho(\pi_\tau)^{-1}$, as if carried directly from the old camera to the new one. 
Keys are deliberately not transformed.
A slot key is an exponential moving average over many writes, so no single writing camera exists to unrotate, and the relative-camera effect the rotation would provide already reaches the read through the values, so the matching stays absolute by design.

\paragraph{VAE design.}
Both autoencoders are frozen before any memory or generator is trained, and within a dataset a single autoencoder serves every latent-space method, so that all reported pixel metrics are measured against a common ceiling.

On \textit{Memory Maze} we finetune convolutional Wan-VAE~\citep{wan2025} and leave its architecture untouched with $16$ latent output channels and causal 3D convolutions giving $8\times$ spatial and $4\times$ temporal compression.
Encoder and decoder are both trained for $40{,}000$ steps on the re-rendered episodes of \cref{app:datasets} according to the training details of \cite{wan2025}.

On \textit{Minecraft} we use MilliVid's released autoencoder~\citep{chandratreya2026millivid} unmodified, encoder and decoder alike.
It is a per-frame vision transformer rather than a convolutional video model over $256\times256$ inputs at patch size $16$, emitting $32$ channels on a $16\times16$ grid, with a four-level pyramid of $256$, $64$, $16$ and $4$ tokens per frame.
It applies \emph{no} temporal compression, so a rollout keeps as many latent frames as it has observations and the memory is written once per frame.
Our memory and generator consume only the finest level of that pyramid with 256 tokens per frame.

\paragraph{DiT design.}
To show that the memory is not tied to one generator, we instantiate it with two diffusion transformers: a camera-conditioned Wan2.1-style model~\citep{wan2025} with $250$M parameters on \textit{Memory Maze} following the model size of~\citet{Po_2025_ICCV} and MilliVid's action-conditioned DiT-B~\citep{peebles2023scalable,chandratreya2026millivid} with $130$M parameters on \textit{Minecraft}.
Both read the memory through channelwise concatenation onto its input channels and every rollout step generates chunks of $N{=}3$ latent frames from a window of $W{=}6$ clean frames.
Everything else follows the respective backbone (\cref{tab:instantiation}).

\textit{Memory Maze.}
The backbone is a stack of $14$ Wan2.1~\citep{wan2025} blocks of width $1{,}024$ with $8$ heads, conditioned only through Wan's per-token timestep modulation, without cross-attention.
A $36$-channel patch embedding maps each cell of the $8\times8$ latent grid to one token, $64$ per frame.
The camera enters attention as PRoPE~\citep{li2025cameras}, a parameter-free camera projection on queries and keys on $40$ head channels, while the remaining carry temporal and 2D-rotary embeddings.
The sequence is laid out as $[\,\textsc{sink} \mid \textsc{context} \mid \textsc{gen}\,]$: frame $0$ is a permanent clean sink token group, the context rows hold the $W$ clean frames before the chunk, and the generation rows hold the noised chunk, which attends to the anchor, its window and itself. Thus, a generation token then attends to at most $(1 + W + N)\times 64 = 640$ tokens.

\textit{Minecraft.}
The backbone is MilliVid's DiT-B~\citep{chandratreya2026millivid}, $12$ blocks of width $768$ with $12$ heads, reproduced parameter for parameter.
Conditioning is per token: every token carries its own adaptive-layer-norm modulation from its noise level, the action embedding (forward, turn left/right) is added to that modulation, and position is an additive sinusoidal code over frame.
Each of the $256$ cells of the $16\times16$ latent grid is one token, embedded by a linear map over $65$ input channels; the $33$ columns beyond the latent are the memory interface and the only parameters absent from the released checkpoint.
Attention is over the $W + N = 9$ frames of a clip, $2{,}304$ tokens against the $3{,}840$ of MilliVid or FramePack, and there is neither camera input nor classifier-free guidance.

\begin{figure}[t]
    \begin{center}
        \centerline{\includegraphics[width=0.75\columnwidth]{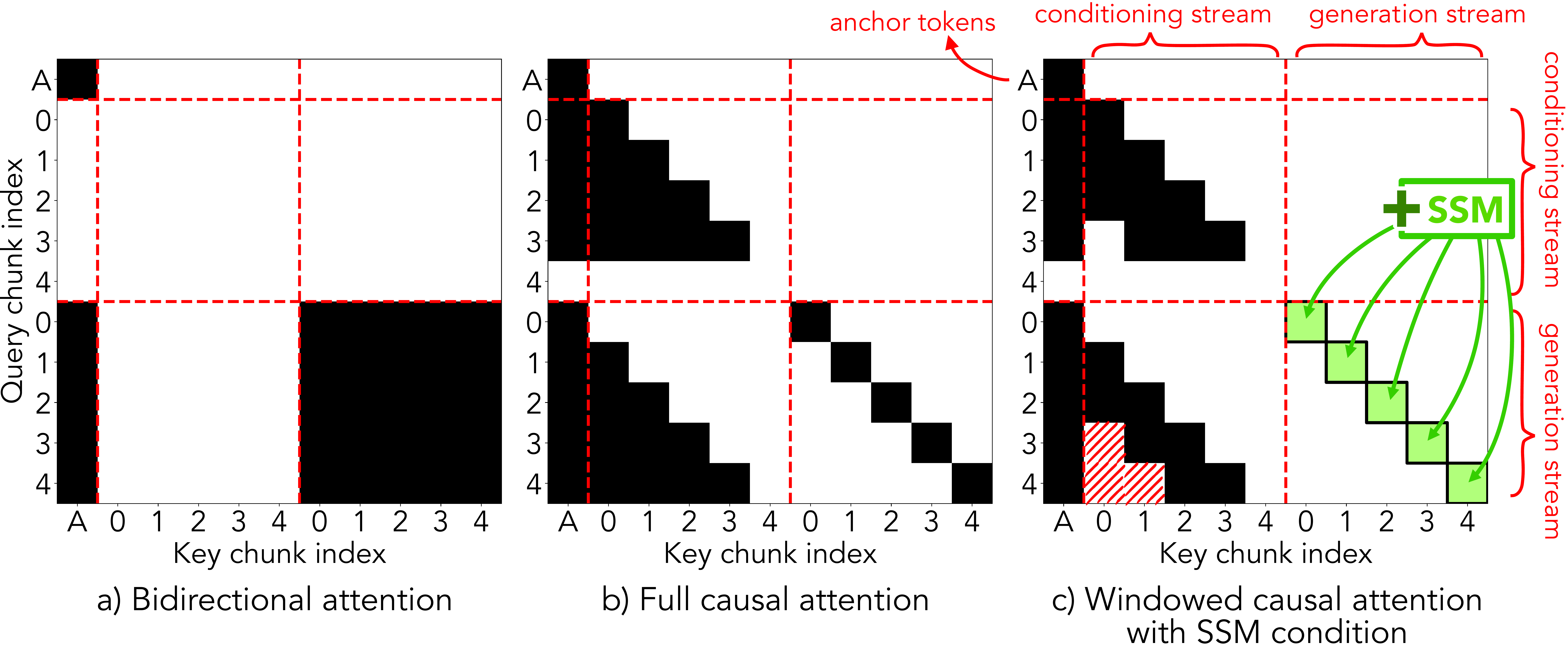}}
        \caption{\textbf{Attention masks.} Allowed attention between query and key chunks for the generators in this work: (a) the bidirectional pretraining of \cref{app:dit-training}, (b) the full-history causal DiT of the Causal~(Full) arm, (c) the windowed chunk-causal attention of \method{}, in which a generation chunk reads the $W$ clean frames before it and the memory's draft. Anchor tokens are the clean frame $0$.}
        \label{fig:attention_masks}
    \end{center}
    \vskip -0.2in
\end{figure}

\begin{table}[t]
    \caption{\textbf{Instantiation of \method{} on the two datasets.} The memory interface, chunk size and context window are shared; everything else follows the backbone each dataset inherits.}
    \label{tab:instantiation}
    \vspace{-2mm}
    \begin{center}
        \begin{scriptsize}
        \setlength{\tabcolsep}{4pt}
        \begin{tabular}{@{}lll@{}}
            \toprule
            & \textit{Memory Maze} & \textit{Minecraft} \\
            \midrule
            Backbone & Wan2.1-style DiT~\citep{wan2025} & DiT-B~\citep{peebles2023scalable,chandratreya2026millivid} \\
            Parameters & $250$M & $130$M \\
            Blocks $\times$ width $\times$ heads & $14 \times 1{,}024 \times 8$ & $12 \times 768 \times 12$ \\
            Control signal & camera pose (PRoPE) & discrete action for modulation \\
            Positional code & temporal and 2D rotary, camera PRoPE & sinusoidal over frame, row and column \\
            Autoencoder & Wan2.1 VAE, finetuned & MilliVid hierarchical autoencoder, pretrained \\
            Frame resolution & $64 \times 64$ & $256 \times 256$ \\
            Latent frame & $16 \times 8 \times 8$, $4\times$ temporal compression & $32 \times 16 \times 16$, per frame \\
            Tokens per frame & $64$ & $256$ \\
            Training clip & $401$ frames & $601$ frames \\
            Chunk $N$ / window $W$ & $3$ / $6$ & $3$ / $6$ \\
            Tokens per denoising step & $640$ & $2{,}304$ \\
            Objective & rectified flow, $v$-prediction & shifted-cosine VP, $v$-prediction \\
            Noise-level draw & logit-normal & uniform \\
            Sampler & $35$ Euler steps & $50$ DDIM steps \\
            Optimizer & AdamW & AdamW \\
            Initialization & causal DiT & causal DiT from~\cite{chandratreya2026millivid} \\
            Memory module & $64.6$M weights, $98{,}304$-float state & $132.8$M weights, $98{,}304$-float state \\
            \bottomrule
        \end{tabular}
        \end{scriptsize}
    \end{center}
    \vspace{-1em}
\end{table}

\paragraph{Memory conditioning.}
The generator reads the memory as a draft rather than through attention.
Before a chunk is denoised, the frozen memory is queried at the pose of each target frame and returns a draft latent with a per-cell confidence $\vc \geq 1$, which becomes the trust value $1 - \tau_c$ with
\begin{equation}
    \tau_c = \mathrm{clip}\!\left(\frac{\alpha/\vc - \epsilon}{\alpha - \epsilon},\; 0,\; \tau_{\max}\right).
    \label{eq:trust}
\end{equation}
The numerator is the expected squared error at the stationary point of \cref{eq:mem-loss}, the denominator its largest value at $\vc = 1$, and $\tau_{\max} = 0.8$.
As in Wan2.1 image-to-video~\citep{wan2025}, the draft is concatenated to the noisy latent along channels as $[\,\text{latent}\,;\,\text{trust}\,;\,\text{draft}\,]$, so that each draft cell is aligned with the cell it predicts and no tokens are added.
Context frames place their own clean latent in the draft slot at trust $1$.
On \textit{Memory Maze} this gives $16 + 4 + 16 = 36$ input channels, the trust being repeated over the $4$ frames folded into each latent frame, and on \textit{Minecraft} $32 + 1 + 32 = 65$, since its autoencoder has no temporal compression.
Once denoised, the chunk is written into the memory for autoregressive generation.

\subsection{Training Formulation}

\paragraph{Spatially-Structured Memory Training.}
\label{app:ssm-training}
The memory is trained alone and before any generator, as a deterministic regressor over posed latents. 
For each training clip ($401$ and $601$ frames), we pre-encode the frames with a pretrained VAE into $101$ and $601$ latent frames for Memory Maze and Minecraft datasets, respectively.
The camera poses are kept at the native frame rate, rebased to $\mE_0 = \mI$, and divided by a single scene-scale constant. 
At every latent position $t$, the model writes the observable prefix $\gO_t$ into its state, is handed one query pose $\pi_{\tau(t)}$, and returns the latents observed for the query pose, $\hat\vz_t = f_\theta(\vz_{\gO_t}, \pi_{\gO_t};\, \pi_{\tau(t)}) \approx \vz_{\tau(t)}$.
Reading operation never modifies the state, so queries are executed at every position in parallel since the recurrence itself trains as a parallel scan, without unrolling~\citep{gsa}.

\textit{Target selection.} 
The target index $\tau(t)$ is drawn independently at each position. 
A \emph{revisit}, $\gC_{\mathrm{rev}}$, is a future frame whose camera lies within $0.5$ world units and a camera direction with cosine similarity larger than $0.9$ that is already in the written prefix, and is answerable only by recalling that visit ($10\%$ of targets).
A \emph{fusion} target, $\gC_{\mathrm{nvs}}$, is a future pose that is \emph{not} a near-duplicate of anything written, but whose viewing footprint is covered by at least one written camera ($45\%$).
It is answerable only by fusing several stored viewpoints.
The \emph{current} frame, $\gC_{\mathrm{cur}}$, reads its own write back out, supervising the write--read round trip end to end ($2\%$).
The remaining $42\%$ recall a uniformly random previously written frame ($\gC_{\mathrm{past}}$) that is re-queried.

\textit{Held-out spans.} 
Independent of the draw above, six blocks of four consecutive latent frames, $24$ of the $101$, are reserved at every step and never written ($\gC_{\mathrm{ho}}$). 
Reserved frames are withheld as revisit evidence and as fusion observers, and each claims two later positions as queries, so about half of all queries ask for a span the memory was never shown and must render from the observations surrounding it.
These positions are scored by a plain squared error rather than the confidence-weighted.

\textit{Objective.} We instantiate \cref{eq:mem-loss} with $\alpha = 0.2$ and $\epsilon = 0.05$.
The two regression terms differ only in $\tau$ and in which state is read:
$\gL^{\mathrm{conf}}_{\mathrm{read}}$ uses the sampled target above, while $\gL^{\mathrm{conf}}_{\mathrm{wf}}$ sets $\tau(t) = t$ and re-reads every frame at its own pose from the frozen final rollout state, which makes it a scene summary.

The address term is $\gL_{\mathrm{addr}} = \lambda_{\mathrm{pose}}\gL_{\mathrm{pose}} + \lambda_{\mathrm{var}}\gL_{\mathrm{var}} + \lambda_{\mathrm{cov}}\gL_{\mathrm{cov}}$.
The anchor asks a per-head \emph{affine} map $(\mW^{p}_{h}, \vb^{p}_{h})$ to recover the viewing geometry of its own tile under each of the $f$ cameras the latent frame folds,
\begin{equation}
    \begin{aligned}
        \gL_{\mathrm{pose}} &= \E_{h,t,j,i}\Big[\tfrac{1}{9}\big\lVert \big(\mW^{p}_{h}\va_{h,t,j} + \vb^{p}_{h}\big)_{i} - \vy_{t,j,i} \big\rVert_2^2\Big], \\
        \vy_{t,j,i} &= \Big[\,\bar\vd_{i,j}\;;\;\big\{(\vp_i + \lambda\,\bar\vd_{i,j})/\sigma\big\}_{\lambda \in \{4,\,16\}}\,\Big] \in \R^{9},
    \end{aligned}
\end{equation}
where $i$ ranges over the $f$ cameras of the fold, $\vp_i$ is the camera centre and $\bar\vd_{i,j}$ the mean unit ray of tile $j$ under camera $i$. 
Collecting the $N{=}BTJ$ addresses of a batch at head $h$ into a centred $\mZ_h \in \R^{N \times d_a}$, one row per address and $d_a$ (the per-head address channels), the spread is the VICReg variance and covariance pair~\citep{bardes2022vicreg},
\begin{equation}
        \gL_{\mathrm{var}} = \E_{h,p}\left[\max\Big(0,\, 1 - \sqrt{\Var(\mZ_h^{p})}\,\Big)\right],
        \quad
        \gL_{\mathrm{cov}} = \E_{h}\left[\tfrac{1}{d_a}\textstyle\sum_{p \neq q}\big[C(\mZ_h)\big]_{pq}^{2}\right],
\end{equation}
where $\mZ_h^{p} \in \R^{N}$ is the $p$-th column of $\mZ_h$ (i.e., coordinate $p$ of every address in the batch) and $C(\mZ_h)$ is its $d_a\times d_a$ covariance.
We use $\lambda_{\mathrm{wf}} = \lambda_{\mathrm{pose}} = 1$, $\lambda_{\mathrm{var}} = 0.1$ and $\lambda_{\mathrm{cov}} = 0.001$.

\textit{Optimization.} 
We use AdamW~\citep{loshchilov2017decoupled} with $\beta = (0.9, 0.95)$, a learning rate of $5e{-4}$ decayed by a cosine schedule, and a weight decay of $0.01$.
Gradient checkpointing is applied to the recurrent blocks, the write-free reads and the camera encoder to keep training feasible on 12 RTX4090 GPUs for $100{,}000$ iterations with an effective batch size of $72$. 

\paragraph{3D Consistent Video Generation Training.}
\label{app:dit-training}
The generator is trained after the memory module is trained according to~\cref{sec:memory,app:ssm-training} and on top of it: 
the memory stays frozen for storing the latent history and to condition the transformer in a causal manner.
It predicts the velocity with one noise level per chunk of $N{=}3$ latent frames and the $W{=}6$ clean frames before the chunk as context, and the two augmentations of \cref{sec:training} are applied at the rates given below.
Half of the clips have each context frame noised with probability $0.5$ to a level drawn from $\gU(0, 0.3]$, and each draft frame is corrupted with probability $0.08$, while the whole draft is replaced by noise for $20\%$ of clips on \textit{Memory Maze} and $5\%$ on \textit{Minecraft}, where a further $15\%$ of clips instead have their context replaced by noise so that the draft is the only source of information.

\textit{Memory Maze DiT training.}
No pretrained generator exists for \textit{Memory Maze}, so we train one in two stages, the second initialized from the first.
(i)~A bidirectional image-to-video model is pretrained on $101$-latent-frame clips with all-to-all attention (\cref{fig:attention_masks}a) and one noise level per clip, for $50{,}000$ steps at an effective batch of $96$ on $12$ RTX4090 GPUs, with a learning rate of $5e{-4}$ and decayed by a cosine schedule.
(ii)~It is then finetuned with the frozen memory: the context is cut to the $W{=}6$ window of \cref{fig:attention_masks}c and the draft and trust channels are attached.
Noise levels are drawn logit-normally per chunk and the target is the rectified-flow velocity.

We use AdamW~\citep{loshchilov2017decoupled} with $\beta = (0.9, 0.95)$, a weight decay of $0.01$, and a learning rate of $1e{-4}$ decayed by a cosine schedule.
The model is finetuned for $15{,}000$ steps at an effective batch of $40$.
At inference every chunk is denoised in $35$ Euler steps and every generated chunk is written back into the memory module.
The Causal~(Full) baseline is a separate finetune of~(i) with the full-history chunk-causal mask of \cref{fig:attention_masks}b, in which every chunk attends to the clean ground-truth history below it, trained for $\approx40{,}000$ steps at batch $72$ with a learning rate of $1e{-4}$ and cosine decay scheduling.

\textit{Minecraft DiT training.}
The generator is initialized from MilliVid's released $7$-frame full-resolution rollout checkpoint, the Causal~(7F) arm of \cref{tab:video-generation-minecraft}.
Only the patch embedding is widened, from $32$ to $65$ input channels for the memory interface.
The target is the $v$-prediction of a shifted-cosine variance-preserving schedule with uniformly drawn noise levels.
Each step loads a full $601$-frame window, draws one boundary $s \sim \gU[6, 598]$, prefills the memory with the ground-truth frames $[0, s)$ and reads the draft of frames $[s, s{+}3)$ from that state without writing, so the draft depends only on frames before the chunk.
The trust label is the same map as for \textit{Memory Maze} with constants, $\alpha = 0.15$ and $\epsilon = 0.007$, and $\tau_{\max} = 0.8$.
We use AdamW~\citep{loshchilov2017decoupled} with a weight decay of $0.01$ and a learning rate of $1.5e{-4}$.
The model is trained for $60{,}000$ steps at an effective batch of $64$ with $4$ accumulation steps.
At inference, the memory is prefilled with the $256$ context frames and each $3$-frame chunk is denoised with $50$-step DDIM sampler and written back into the memory.

\subsection{3D Consistent Video Generation Evaluation} \label{app:protocols}

\paragraph{Memory design.}
The \method{}-SM is instantiated identically on both datasets, as described in \cref{sec:memory} and \cref{app:impl} with $8$ layers of $6$ heads, head dimension $64$, $M{=}16$ slots per head, top-$1$ routing whose sigmoid write amplitude is left unnormalized, a write gate tied to the Mamba-2 decay, the SE(3) value warp acting on all head channels as detailed in \cref{par:value-transport}.
Keys and queries are functions of pose alone, and each frame contributes $16$ write tokens.
The state is therefore $2\times L\times H\times M\times d = 98{,}304$ floats on both datasets and at every horizon.

\paragraph{Baseline Implementation Details.}
Every baseline is generated based on its authors' implementation and scored by us after adapting for the benchmarks.
On \textit{Memory Maze}, the GeoCache (based on Lyra2.0~\citep{shen2026lyra}) and DFoT~\citep{song2025historyguided} architecture, diffusion backend, sampler, sliding-window schedule and guidance belong to the reference implementation, and evaluated on our test dataset after training models from scratch.
On \textit{Minecraft}, the rollout is produced by the released code unchanged. All methods are run on the same episodes and decoded by a shared autoencoder.

\textit{Causal DiT.}
On \textit{Memory Maze}, Causal~(Full) is the full-history causal finetune of our bidirectional model (\cref{app:dit-training}), scored as an arm of its own with the same $35$-step rollout and key--value cache, and Causal~(100F) is the same recipe with the clean context capped at $25$ latent frames ($100$ pixel frames).
On \textit{Minecraft}, Causal~(7F) is MilliVid's released full-resolution rollout checkpoint, run unmodified.

\textit{DFoT.}
\citet{song2025historyguided} assigns an independent noise level to every token and denoises a sliding window under history guidance.
We train it on the latents, clips and camera convention of \cref{app:datasets}: $13$ layers, width $1024$, $8$ heads and patch size $1$ over the $8\times8$ latent grid, giving $249$M parameters against the $250$M budget of our own generator, on $101$-frame latent clips.
To achieve a fair comparison, we introduce two departures from the released code:
(i) Camera conditioning is PRoPE~\citep{li2025cameras}, which applies a parameter-free rotation of queries and keys, replacing the reference model's ray-encoding pathway, 
(ii) the noise-level schedule draws the context--horizon split per sample rather than at a fixed position, so every window the sampler later requests is in distribution, and
History guidance remains unmodified and the model is optimized with AdamW~\citep{loshchilov2017decoupled} at $\mathrm{lr}=1e{-4}$ at an effective batch size of $96$ for $175{,}000$ steps in total.
We use the reference sampler at $50$ steps with stabilized history guidance at scale $4.0$.
Each window represents $50$ latent frames and generates $51$, so DFoT carries a bounded $50$-frame history rather than the whole trajectory equivalent to $\approx200$ pixel frames which matches our training samples on average.

\textit{GeoCache.}
GeoCache is our reimplementation of the geometry-cached memory of Lyra~2.0~\citep{shen2026lyra}, strictly following its official repository: every observed frame is cached with its depth and camera, a read reprojects the cached geometry into the target camera to greedily retrieve the frames covering the most new target pixels, and use them with FramePack-style context~\citep{zhang2025framepack} both as latent slots and as depth-warped correspondence maps.
We train it with the official codebase and recipe, including its self-augmentation against exposure bias, and change only what the maze setting requires.

The backbone is a pretrained $14$-layer bidirectional model without cross-attention ($292$M parameters with Lyra's camera and correspondence encoders), as the released $14$B system is untrainable at this scale, and each step attends an anchor, six temporal and nine spatial latent frames around a three-frame chunk, uncompressed to match our generator's window.
For $64\times64$ frames the cache keeps full-resolution depth at every frame, and pixels without valid depth are excluded from retrieval.
So that copying from the cache is learnable, training retrieves only from frames preceding the chunk, for three target views as at inference, without corrupting or dropping the retrieved slots; self-augmentation is widened to the full noise range with four denoising steps, since the released setting only presents near-clean history.
The model is trained for about $78{,}000$ iterations with a learning rate of $1e{-4}$, and a decaying learning rate schedule.
We evaluate it with simulator depth, the most favourable setting for an explicit cache, starting the return leg from the last observed frame and a cache seeded with the whole forward leg and $35$ denoising steps per chunk.

\textit{FramePack.}
\cite{zhang2025framepack} compresses history by representing older frames with progressively fewer tokens.
We evaluate the released checkpoint of the reimplementation distributed with MilliVid~\citep{chandratreya2026millivid} to share its backbone, autoencoder, and sampler with other \textit{Minecraft} arms.
The progressive compression on subsequent $192$, $48$, $12$, and the most recent $3$ frame groups add up to $255$ frames of history and occupies $3{,}840$-token budget with $32$-channel tokens.
Three frames are emitted per step and is run at $50$ sampling steps.

\textit{MilliVid.}
\cite{chandratreya2026millivid} conditions on a learned coarse-to-fine latent pyramid rather than a spatial repatchification, encoding each frame at four levels of $256$, $64$, $16$ and $4$ tokens and spending an equal token budget on each.
Its history likewise spans $255$ frames, and a chunk advances via $85$ rollout steps per $192$-frame chunk equivalent to $85\times50=4{,}250$ denoising steps.
Both it and FramePack use the authors' released checkpoints with a DiT-B backbone~\citep{peebles2023scalable}, the same size as our own \textit{Minecraft} generator and no classifier-free guidance at test time.

\paragraph{Validating the state-parameter count.} For our comparisons in~\cref{tab:video-generation-minecraft}, we keep a matched memory/context size for FramePack~\citep{zhang2025framepack}, MilliVid~\citep{chandratreya2026millivid}, and our approach for the long horizon memory.

\textit{FramePack and MilliVid.} 
Uses $3{,}072$ tokens for context (which include enough budget for [3, 12, 48, 192] frames at levels 0-3 respectively) and $768$ tokens reserved for prediction in any given step. 
At inference time, the model sees a maximum of $3 + 12 + 48 + 192 = 255$ frames into the past.
$3{,}072$ tokens add up to $3{,}072\times32=98{,}304$ floats.

In \method{}'s $8$ layer, $6$ head per layer, $16$ memory slot per head, $64$ dimensional memory module,
$2\times8\times6\times16\times64=98{,}304\;\text{floats}$ are essentially providing context for the diffusion process in an unbounded horizon.
Although we validate that expanding the fixed number of memory slots also improves the reconstruction quality of our memory module (see~\cref{tab:expand-memory}), we use our $98{,}304$ float memory module throughout this paper for fair comparison.

\begin{table}[t]
    \caption{\textbf{Memory capacity.} Reconstruction quality of the memory alone on \textit{Memory Maze}, trained on $401$-frame episodes.
    Doubling the memory parameters gains $0.3$\,dB at $401$ frames and $1.3$\,dB at $3{,}201$.}
    \label{tab:expand-memory}
    \vspace{-2mm}
    \begin{center}
        \begin{scriptsize}
        \begin{sc}
          \setlength{\tabcolsep}{4pt}
          \begin{tabular}{@{}lccccccccc@{}}
            \multirow{2}{*}{Variant} & \multicolumn{3}{c}{\normalfont\itshape 401 frames} & \multicolumn{3}{c}{\normalfont\itshape 1601 frames} & \multicolumn{3}{c}{\normalfont\itshape 3201 frames} \\
            \cmidrule(lr){2-4} \cmidrule(lr){5-7} \cmidrule(lr){8-10}
            & \metric{PSNR$\uparrow$} & \metric{SSIM$\uparrow$} & \metric{LPIPS$\downarrow$} & \metric{PSNR$\uparrow$} & \metric{SSIM$\uparrow$} & \metric{LPIPS$\downarrow$} & \metric{PSNR$\uparrow$} & \metric{SSIM$\uparrow$} & \metric{LPIPS$\downarrow$} \\
            \midrule
            \method{}-Slot Memory & 32.7 & 0.940 & 0.024 & 32.4 & 0.936 & 0.024 & 31.9 & 0.931 & 0.026 \\
            \cmidrule(lr){1-10}
            Scaling Up Memory Slots $(\times2)$ & \textbf{33.0} & \textbf{0.943} & \textbf{0.023} & \textbf{33.3} & \textbf{0.947} & \textbf{0.020} & \textbf{33.2} & \textbf{0.946} & \textbf{0.020} \\
            \bottomrule
          \end{tabular}%
        \end{sc}
        \end{scriptsize}
        \vspace{-1mm}
    \end{center}
\end{table}

\begin{table}[t]
    \caption{\textbf{Comparison against StateSpaceDiffuser~\citep{savov2026statespacediffuser} reimplementation on \textit{Memory Maze}.} Both generators share the DiT architecture and draft conditioning differing in how the memory is conditioned and read. Column groups are total trajectory length. PSNR and SSIM are computed over each target frame that were visible during the forward leg and averaged per frame.}
    \label{tab:action-memory}
    \vspace{-1mm}
    \begin{center}
        \begin{scriptsize}
        \begin{sc}
          \setlength{\tabcolsep}{4pt}
          \begin{tabular}{@{}lccccccccc@{}}
            \multirow{2}{*}{Method} & \multicolumn{3}{c}{\normalfont\itshape $(25 + 21) + 25 = 71$ frames} & \multicolumn{3}{c}{\normalfont\itshape $(51 + 21) + 51 = 123$ frames} & \multicolumn{3}{c}{\normalfont\itshape $(101 + 21) + 101 = 223$ frames} \\
            \cmidrule(lr){2-4} \cmidrule(lr){5-7} \cmidrule(lr){8-10}
            & \metric{PSNR$\uparrow$} & \metric{SSIM$\uparrow$} & \metric{LPIPS$\downarrow$} & \metric{PSNR$\uparrow$} & \metric{SSIM$\uparrow$} & \metric{LPIPS$\downarrow$} & \metric{PSNR$\uparrow$} & \metric{SSIM$\uparrow$} & \metric{LPIPS$\downarrow$} \\
            \midrule
            StateSpaceDiffuser & 28.5 & 0.874 & 0.145 & 24.8 & 0.806 & 0.177 & 21.0 & 0.717 & 0.290 \\
            \method{} & \textbf{30.2} & \textbf{0.906} & \textbf{0.108} & \textbf{32.3} & \textbf{0.931} & \textbf{0.073} & \textbf{33.3} & \textbf{0.939} & \textbf{0.039} \\
            \bottomrule
          \end{tabular}
        \end{sc}
        \end{scriptsize}
        \vspace{-2mm}
    \end{center}
\end{table}

\subsection{Additional Experiments.} \label{app:additional-experiments}

\paragraph{Qualitative results for Memory Maze.}
As reflected by the numbers in \cref{tab:video-generation}, \method{} preserves the geometry, layout, and appearance of previously observed regions most consistently over the full return trajectory. In contrast, the baselines increasingly exhibit geometric drift, misplaced structures, and appearance changes as the revisit horizon grows, with the degradation becoming particularly visible in the later frame ranges.

\definecolor{rowgray}{RGB}{58,58,58}
\definecolor{panelgray}{RGB}{232,232,232}

\newcommand{\imageRoot}{Figures/raw_pngs/experiments/ablations/maze-qualitative-generation}

\newcommand{\panel}[2]{%
  \includegraphics[
    width=1.9cm,
    height=1.9cm,
    keepaspectratio=false
  ]{#1}%
}

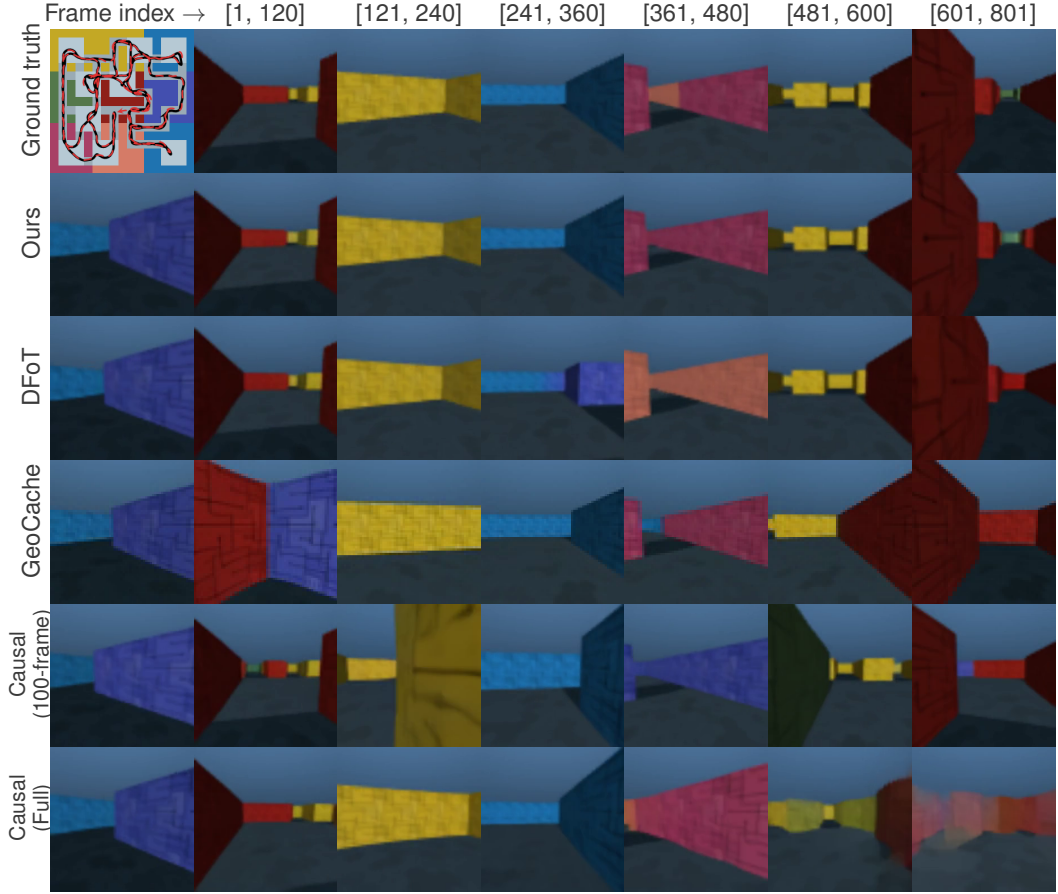
\begin{figure}[t]
  \centering

  \begin{tikzpicture}[x=1cm,y=1cm]
    \def\colw{1.9}
    \def\rowh{1.9}
    \def\leftpad{0}

    \foreach \col/\range [count=\i from 1] in {
      1/{[1, 120]},
      2/{[121, 240]},
      3/{[241, 360]},
      4/{[361, 480]},
      5/{[481, 600]},
      6/{[601, 801]}
    }
    {
      \node[
        font=\small\sffamily,
        rowgray,
        anchor=south,
        inner sep=0pt
      ] at ({\leftpad+(\i+0.5)*\colw},0.04) {\range};
    }

    \node[
      font=\small\sffamily,
      rowgray,
      anchor=south west,
      inner sep=0pt
    ] at (-0.08,0.10) {Frame index $\to$};

    \foreach \rowfolder/\files [count=\r from 0] in {
      row1/{01_topdown.png,02_Image25.png,03_Image26.png,04_Image27.png,05_Image28.png,06_Image29.png,07_Image30.png},
      row2/{01_Image17.png,02_Image18.png,03_Image19.png,04_Image20.png,05_Image21.png,06_Image22.png,07_Image23.png},
      row3/{01_Image9.png,02_Image10.png,03_Image11.png,04_Image12.png,05_Image13.png,06_Image14.png,07_Image15.png},
      row4/{01_Image1.png,02_Image2.png,03_Image3.png,04_Image4.png,05_Image5.png,06_Image6.png,07_Image7.png},
      row5/{01_Image31.png,02_Image32.png,03_Image33.png,04_Image34.png,05_Image35.png,06_Image36.png,07_Image37.png},
      row6/{01_Image38.png,02_Image39.png,03_Image40.png,04_Image41.png,05_Image42.png,06_Image43.png,07_Image44.png},
    }
    {
      \ifnum\r=4
        \node[
          rotate=90,
          anchor=center,
          font=\fontsize{8}{5}\selectfont\sffamily,
          rowgray
        ] at (-0.25,{-0.8-\r*\rowh})
          {\shortstack{Causal\\(100-frame)}};
      \else\ifnum\r=5
        \node[
          rotate=90,
          anchor=center,
          font=\fontsize{8}{5}\selectfont\sffamily,
          rowgray
        ] at (-0.25,{-0.8-\r*\rowh})
          {\shortstack{Causal\\(Full)}};
      \else
        \node[
          rotate=90,
          anchor=center,
          font=\small\sffamily,
          rowgray
        ] at (-0.25,{-0.8-\r*\rowh})
          {\ifcase\r
            Ground truth
            \or Ours
            \or DFoT
            \or GeoCache
          \fi};
      \fi\fi

      \foreach \file [count=\col from 1] in \files
      {
        \pgfmathtruncatemacro{\x}{\col-1}
        \pgfmathtruncatemacro{\y}{\r}

        \node[
          inner sep=0pt,
          anchor=north west
        ] at ({\leftpad+\x*\colw},{-\y*\rowh})
        {\panel{\imageRoot/\rowfolder/\file}{panel-\r-\col}};
      }
    }
  \end{tikzpicture}
  \vspace{-3em}
  \caption{\textbf{Qualitative comparison on Memory Maze.} Generated frames for each method in~\cref{tab:video-generation}, on the same trajectory with $823$ frame context frame and $801$ target frame generation.}
  \label{fig:qualitative-memory-maze}
\end{figure}

\paragraph{Comparison with action-conditioned memory.}
Additionally, we demonstrate that pose-conditioned memory can enable
longer-horizon video generation with a comparison against StateSpaceDiffuser~\citep{savov2026statespacediffuser}
which conditions its state-space memory on the agent's actions rather than querying it by camera pose.
We reimplement it on \textit{Memory Maze} inside our generator for the same number of training iterations for the SSM and the DiT. 
The DiT, chunk size, context window and draft conditioning match \method{}, but the memory is a Mamba2~\citep{mamba2} recurrence driven by actions, and the DiT receives actions through its timestep modulation instead of the camera.
\Cref{tab:action-memory} compares both on trajectories with forward legs of $25$, $51$, and $101$ frames.
Throughout the episodes, we observe that the action-conditioned memory degrades steadily with the horizon, losing $7.5$\,dB of PSNR on seen regions while \method{} improves as more of the maze is observed.
Because only regions observed during the forward leg are scored, this is a failure to recall rather than to invent, and we attribute it to the fact that the memory has no query: 
it predicts the next frame from the frames and actions so far. Recalling a place seen earlier requires the recurrence to integrate every action taken since that visit, 
and errors in this implicit reckoning grow with the length of the path.

\definecolor{qgrowgray}{RGB}{58,58,58}

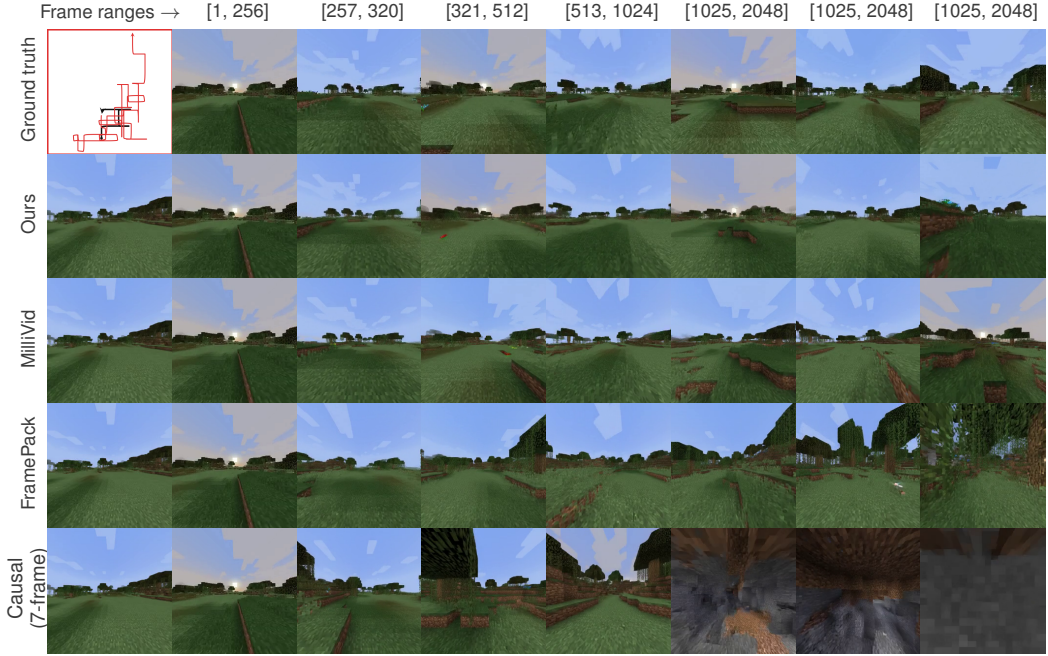
\begin{figure}[t]
  \centering

\newcommand{\qqgImageRoot}{Figures/raw_pngs/experiments/227}

\newcommand{\qqgPanel}[1]{%
  \includegraphics[
    width=1.65cm,
    height=1.65cm,
    keepaspectratio=false
  ]{#1}%
}

  \begin{tikzpicture}[x=1cm,y=1cm]
    \def\colw{1.65}
    \def\rowh{1.65}
    \def\leftpad{0}

    \foreach \range [count=\i from 1] in {
      {[1, 256]},
      {[257, 320]},
      {[321, 512]},
      {[513, 1024]},
      {[1025, 2048]},
      {[1025, 2048]},
      {[1025, 2048]}
    }
    {
      \node[
        font=\scriptsize\sffamily,
        qgrowgray,
        anchor=south,
        inner sep=0pt,
        xshift=0.05cm,
        yshift=0.06cm
      ] at ({\leftpad+(\i+0.5)*\colw},0.04) {\range};
    }

    \node[
      font=\scriptsize\sffamily,
      qgrowgray,
      anchor=south west,
      inner sep=0pt
    ] at (-0.1,0.08) {Frame ranges $\to$};

    \foreach \rowfolder [count=\r from 0] in {row1,row2,row3,row4,row5}{
      \ifcase\r
        \def\files{01_topdown.png,02_Image34.png,03_Image35.png,04_Image36.png,05_Image37.png,06_Image38.png,07_Image39.png,08_Image40.png}
      \or
        \def\files{01_Image25.png,02_Image26.png,03_Image27.png,04_Image28.png,05_Image29.png,06_Image30.png,07_Image31.png,08_Image32.png}
      \or
        \def\files{01_Image17.png,02_Image18.png,03_Image19.png,04_Image20.png,05_Image21.png,06_Image22.png,07_Image23.png,08_Image24.png}
      \or
        \def\files{01_Image9.png,02_Image10.png,03_Image11.png,04_Image12.png,05_Image13.png,06_Image14.png,07_Image15.png,08_Image16.png}
      \or
        \def\files{01_Image1.png,02_Image2.png,03_Image3.png,04_Image4.png,05_Image5.png,06_Image6.png,07_Image7.png,08_Image8.png}
      \fi

      \ifnum\r=0
        \node[
          rotate=90,
          anchor=center,
          font=\scriptsize\sffamily,
          qgrowgray
        ] at (-0.25,{-0.8-\r*\rowh}) {Ground truth};
      \else\ifnum\r=1
        \node[
          rotate=90,
          anchor=center,
          font=\scriptsize\sffamily,
          qgrowgray
        ] at (-0.25,{-0.8-\r*\rowh}) {Ours};
      \else\ifnum\r=2
        \node[
          rotate=90,
          anchor=center,
          font=\scriptsize\sffamily,
          qgrowgray
        ] at (-0.25,{-0.8-\r*\rowh}) {MilliVid};
      \else\ifnum\r=3
        \node[
          rotate=90,
          anchor=center,
          font=\scriptsize\sffamily,
          qgrowgray
        ] at (-0.25,{-0.8-\r*\rowh}) {FramePack};
      \else
        \node[
          rotate=90,
          anchor=center,
          font=\fontsize{7.5}{5}\selectfont\sffamily,
          qgrowgray
        ] at (-0.25,{-0.80-\r*\rowh})
          {\shortstack{Causal\\(7-frame)}};
      \fi\fi\fi\fi

      \foreach \file [count=\col from 1] in \files{
        \pgfmathtruncatemacro{\x}{\col-1}
        \pgfmathtruncatemacro{\y}{\r}

        \node[
          inner sep=0pt,
          anchor=north west
        ] at ({\leftpad+\x*\colw},{-\y*\rowh})
        {\qqgPanel{\qqgImageRoot/\rowfolder/\file}};
      }
    }
  \end{tikzpicture}

  \caption{\textbf{Additional qualitative comparisons on \textit{Minecraft}.} Generated frames
    for each method in \cref{tab:video-generation-minecraft}, on the same
    trajectory, at the 2048 frame horizon with 256 initial context frames. Note that only \method{} faithfully retrieves the position of the sun. }
  \label{fig:qualitative-minecraft-227}
\end{figure}

\definecolor{qgrowgray}{RGB}{58,58,58}

\begin{figure}[t]
  \centering

\newcommand{\qqqqgImageRoot}{Figures/raw_pngs/experiments/132}

\newcommand{\qqqqgPanel}[1]{%
  \includegraphics[
    width=1.65cm,
    height=1.65cm,
    keepaspectratio=false
  ]{#1}%
}

  \begin{tikzpicture}[x=1cm,y=1cm]
    \def\colw{1.65}
    \def\rowh{1.65}
    \def\leftpad{0}

    \foreach \range [count=\i from 1] in {
      {[1, 256]},
      {[257, 320]},
      {[321, 512]},
      {[513, 1024]},
      {[1025, 2048]},
      {[1025, 2048]},
      {[1025, 2048]}
    }
    {
      \node[
        font=\scriptsize\sffamily,
        qgrowgray,
        anchor=south,
        inner sep=0pt,
        xshift=0.05cm,
        yshift=0.06cm
      ] at ({\leftpad+(\i+0.5)*\colw},0.04) {\range};
    }

    \node[
      font=\scriptsize\sffamily,
      qgrowgray,
      anchor=south west,
      inner sep=0pt
    ] at (-0.1,0.08) {Frame ranges $\to$};

    \foreach \rowfolder [count=\r from 0] in {row1,row2,row3,row4,row5}{
      \ifcase\r
        \def\files{01_topdown.png,02_Image34.png,03_Image35.png,04_Image36.png,05_Image37.png,06_Image38.png,07_Image39.png,08_Image40.png}
      \or
        \def\files{01_Image25.png,02_Image26.png,03_Image27.png,04_Image28.png,05_Image29.png,06_Image30.png,07_Image31.png,08_Image32.png}
      \or
        \def\files{01_Image17.png,02_Image18.png,03_Image19.png,04_Image20.png,05_Image21.png,06_Image22.png,07_Image23.png,08_Image24.png}
      \or
        \def\files{01_Image9.png,02_Image10.png,03_Image11.png,04_Image12.png,05_Image13.png,06_Image14.png,07_Image15.png,08_Image16.png}
      \or
        \def\files{01_Image1.png,02_Image2.png,03_Image3.png,04_Image4.png,05_Image5.png,06_Image6.png,07_Image7.png,08_Image8.png}
      \fi

      \ifnum\r=0
        \node[
          rotate=90,
          anchor=center,
          font=\scriptsize\sffamily,
          qgrowgray
        ] at (-0.25,{-0.8-\r*\rowh}) {Ground truth};
      \else\ifnum\r=1
        \node[
          rotate=90,
          anchor=center,
          font=\scriptsize\sffamily,
          qgrowgray
        ] at (-0.25,{-0.8-\r*\rowh}) {Ours};
      \else\ifnum\r=2
        \node[
          rotate=90,
          anchor=center,
          font=\scriptsize\sffamily,
          qgrowgray
        ] at (-0.25,{-0.8-\r*\rowh}) {MilliVid};
      \else\ifnum\r=3
        \node[
          rotate=90,
          anchor=center,
          font=\scriptsize\sffamily,
          qgrowgray
        ] at (-0.25,{-0.8-\r*\rowh}) {FramePack};
      \else
        \node[
          rotate=90,
          anchor=center,
          font=\fontsize{7.5}{5}\selectfont\sffamily,
          qgrowgray
        ] at (-0.25,{-0.80-\r*\rowh})
          {\shortstack{Causal\\(7-frame)}};
      \fi\fi\fi\fi

      \foreach \file [count=\col from 1] in \files{
        \pgfmathtruncatemacro{\x}{\col-1}
        \pgfmathtruncatemacro{\y}{\r}

        \node[
          inner sep=0pt,
          anchor=north west
        ] at ({\leftpad+\x*\colw},{-\y*\rowh})
        {\qqqqgPanel{\qqqqgImageRoot/\rowfolder/\file}};
      }
    }
  \end{tikzpicture}

  \caption{\textbf{Additional qualitative comparisons on \textit{Minecraft}.}}
  \label{fig:qualitative-minecraft-132}
\end{figure}
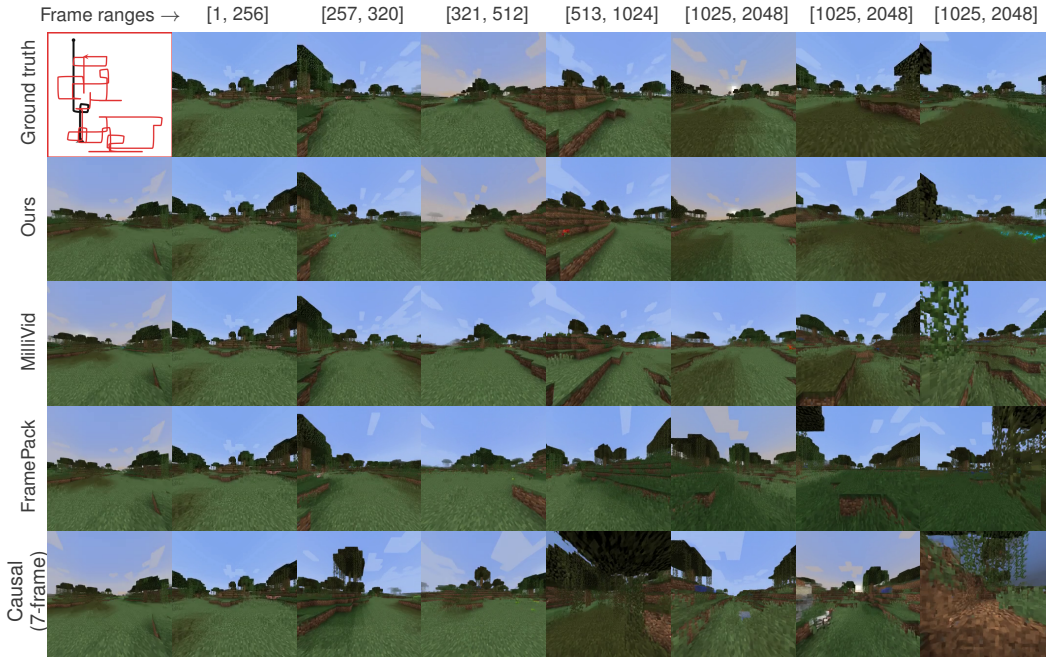

\definecolor{qgrowgray}{RGB}{58,58,58}

\newcommand{\qgImageRoot}{Figures/raw_pngs/experiments/qualitative-generation}

\newcommand{\qgPanel}[1]{%
  \includegraphics[
    width=1.65cm,
    height=1.65cm,
    keepaspectratio=false
  ]{#1}%
}

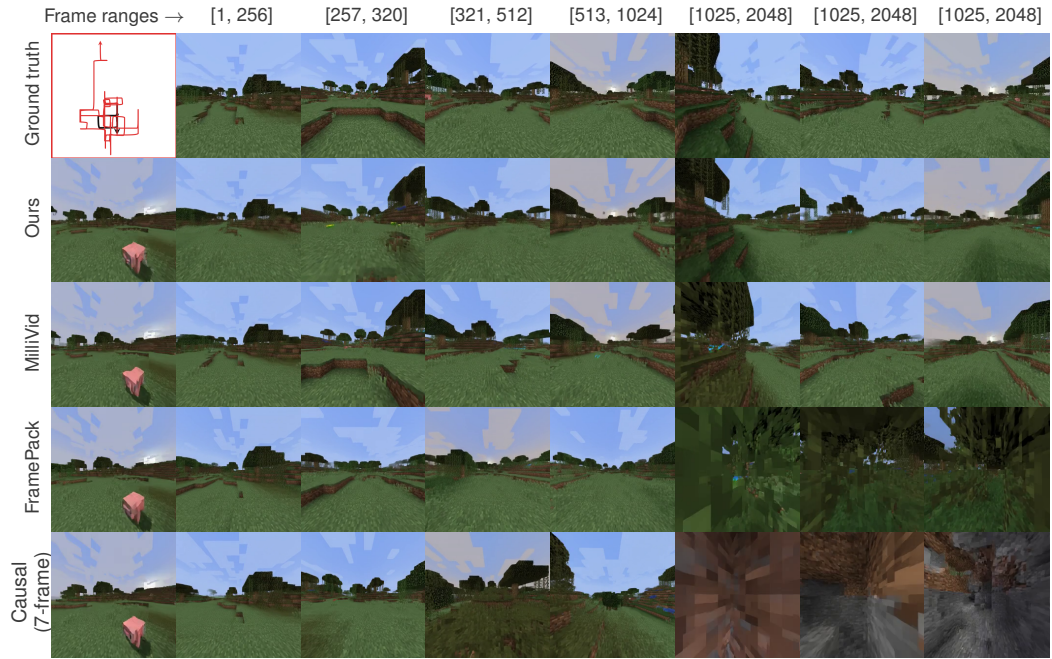
\begin{figure}[t]
  \centering

  \begin{tikzpicture}[x=1cm,y=1cm]
    \def\colw{1.65}
    \def\rowh{1.65}
    \def\leftpad{0}

    \foreach \range [count=\i from 1] in {
      {[1, 256]},
      {[257, 320]},
      {[321, 512]},
      {[513, 1024]},
      {[1025, 2048]},
      {[1025, 2048]},
      {[1025, 2048]}
    }
    {
      \node[
        font=\scriptsize\sffamily,
        qgrowgray,
        anchor=south,
        inner sep=0pt,
        xshift=0.05cm,
        yshift=0.06cm
      ] at ({\leftpad+(\i+0.5)*\colw},0.04) {\range};
    }

    \node[
      font=\scriptsize\sffamily,
      qgrowgray,
      anchor=south west,
      inner sep=0pt
    ] at (-0.1,0.08) {Frame ranges $\to$};

    \foreach \rowfolder [count=\r from 0] in {row1,row2,row3,row4,row5}{
      \ifcase\r
        \def\files{01_topdown.png,02_Image34.png,03_Image35.png,04_Image36.png,05_Image37.png,06_Image38.png,07_Image39.png,08_Image40.png}
      \or
        \def\files{01_Image25.png,02_Image26.png,03_Image27.png,04_Image28.png,05_Image29.png,06_Image30.png,07_Image31.png,08_Image32.png}
      \or
        \def\files{01_Image17.png,02_Image18.png,03_Image19.png,04_Image20.png,05_Image21.png,06_Image22.png,07_Image23.png,08_Image24.png}
      \or
        \def\files{01_Image9.png,02_Image10.png,03_Image11.png,04_Image12.png,05_Image13.png,06_Image14.png,07_Image15.png,08_Image16.png}
      \or
        \def\files{01_Image1.png,02_Image2.png,03_Image3.png,04_Image4.png,05_Image5.png,06_Image6.png,07_Image7.png,08_Image8.png}
      \fi

      \ifnum\r=0
        \node[
          rotate=90,
          anchor=center,
          font=\scriptsize\sffamily,
          qgrowgray
        ] at (-0.25,{-0.8-\r*\rowh}) {Ground truth};
      \else\ifnum\r=1
        \node[
          rotate=90,
          anchor=center,
          font=\scriptsize\sffamily,
          qgrowgray
        ] at (-0.25,{-0.8-\r*\rowh}) {Ours};
      \else\ifnum\r=2
        \node[
          rotate=90,
          anchor=center,
          font=\scriptsize\sffamily,
          qgrowgray
        ] at (-0.25,{-0.8-\r*\rowh}) {MilliVid};
      \else\ifnum\r=3
        \node[
          rotate=90,
          anchor=center,
          font=\scriptsize\sffamily,
          qgrowgray
        ] at (-0.25,{-0.8-\r*\rowh}) {FramePack};
      \else
        \node[
          rotate=90,
          anchor=center,
          font=\fontsize{7.5}{5}\selectfont\sffamily,
          qgrowgray
        ] at (-0.25,{-0.80-\r*\rowh})
          {\shortstack{Causal\\(7-frame)}};
      \fi\fi\fi\fi

      \foreach \file [count=\col from 1] in \files{
        \pgfmathtruncatemacro{\x}{\col-1}
        \pgfmathtruncatemacro{\y}{\r}

        \node[
          inner sep=0pt,
          anchor=north west
        ] at ({\leftpad+\x*\colw},{-\y*\rowh})
        {\qgPanel{\qgImageRoot/\rowfolder/\file}};
      }
    }
  \end{tikzpicture}

  \caption{\textbf{Additional qualitative comparisons on \textit{Minecraft}.}}
  \label{fig:qualitative-minecraft-appendix}
\end{figure}

\end{document}